\documentclass[journal]{IEEEtran}

\usepackage[utf8]{inputenc}
\usepackage[pdftex]{graphicx}
\usepackage{amsmath}
\usepackage{amssymb}
\usepackage{algorithmic}
\usepackage{algorithm}
\usepackage[english]{babel}
\usepackage{color}
\usepackage{multicol}
\usepackage{multirow}
\usepackage{tabularx,colortbl}
\usepackage[compress]{cite}
\usepackage{subfigure}
\usepackage{balance}
\usepackage{mathtools}

\newcommand{\argmax}{\operatornamewithlimits{arg\ max}}

\begin{document}

\title{One-class classifiers based on entropic spanning graphs}

\author{Lorenzo Livi,~\IEEEmembership{Member,~IEEE},
       and Cesare Alippi,~\IEEEmembership{Fellow,~IEEE}
\thanks{Manuscript received ; revised .}
\thanks{Lorenzo Livi and Cesare Alippi are with the Dept. of Electronics, Information, and Bioengineering, Politecnico di Milano, Piazza Leonardo da Vinci 32, 20133 Milano, Italy and ALaRI, Faculty of Informatics, Universit\`a della Svizzera Italiana, Via Giuseppe Buffi 13, 6904 Lugano, Switzerland (e-mail: \{lorenzo.livi, cesare.alippi\}@polimi.it).}}

\maketitle

\begin{abstract}
One-class classifiers offer valuable tools to assess the presence of outliers in data.
In this paper, we propose a design methodology for one-class classifiers based on entropic spanning graphs.
Our approach takes into account the possibility to process also non-numeric data by means of an embedding procedure.
The spanning graph is learned on the embedded input data and the outcoming partition of vertices defines the classifier.
The final partition is derived by exploiting a criterion based on mutual information minimization.
Here, we compute the mutual information by using a convenient formulation provided in terms of the $\alpha$-Jensen difference.
Once training is completed, in order to associate a confidence level with the classifier decision, a graph-based fuzzy model is constructed.
The fuzzification process is based only on topological information of the vertices of the entropic spanning graph.
As such, the proposed one-class classifier is suitable also for data characterized by complex geometric structures.
We provide experiments on well-known benchmarks containing both feature vectors and labeled graphs.
In addition, we apply the method to the protein solubility recognition problem by considering several representations for the input samples.
Experimental results demonstrate the effectiveness and versatility of the proposed method with respect to other state-of-the-art approaches.
\end{abstract}

\section{Introduction}
\label{sec:intro}

The analysis of large volumes of data is hampered by many technical problems, including the ones related to the quality and interpretation of associated information.
One-class classifier design is an important research endeavour \cite{6857384,dufrenois2016one} that can be used to tackle problems of anomaly/novelty detection or, more generally, to recognize outliers in incoming data \cite{ecoli_graph_complexity,7038137,alippi_hierarchical_tnnls2016,6889860,radovanovic2015reverse,abdallah2016anynovel}.
Several different methods have been proposed in the literature, including clustering-based techniques, kernel methods, and statistical approaches (see \cite{pimentel2014review} for a recent survey).
A widespread approach considers data generating processes whose ``nominal conditions'' are known by field experts, while non-nominal conditions are unknown; the fault recognition problem provides a pertinent example in this direction \cite{occ_sg_enricods}.
In such cases, a one-class classifier can be effectively trained to recognize the nominal conditions only and reject, along with a confidence level, instances lying outside the nominal class.

\begin{figure}[ht!]
\centering

\subfigure[Embedding.]{
\includegraphics[scale=0.4,keepaspectratio=true]{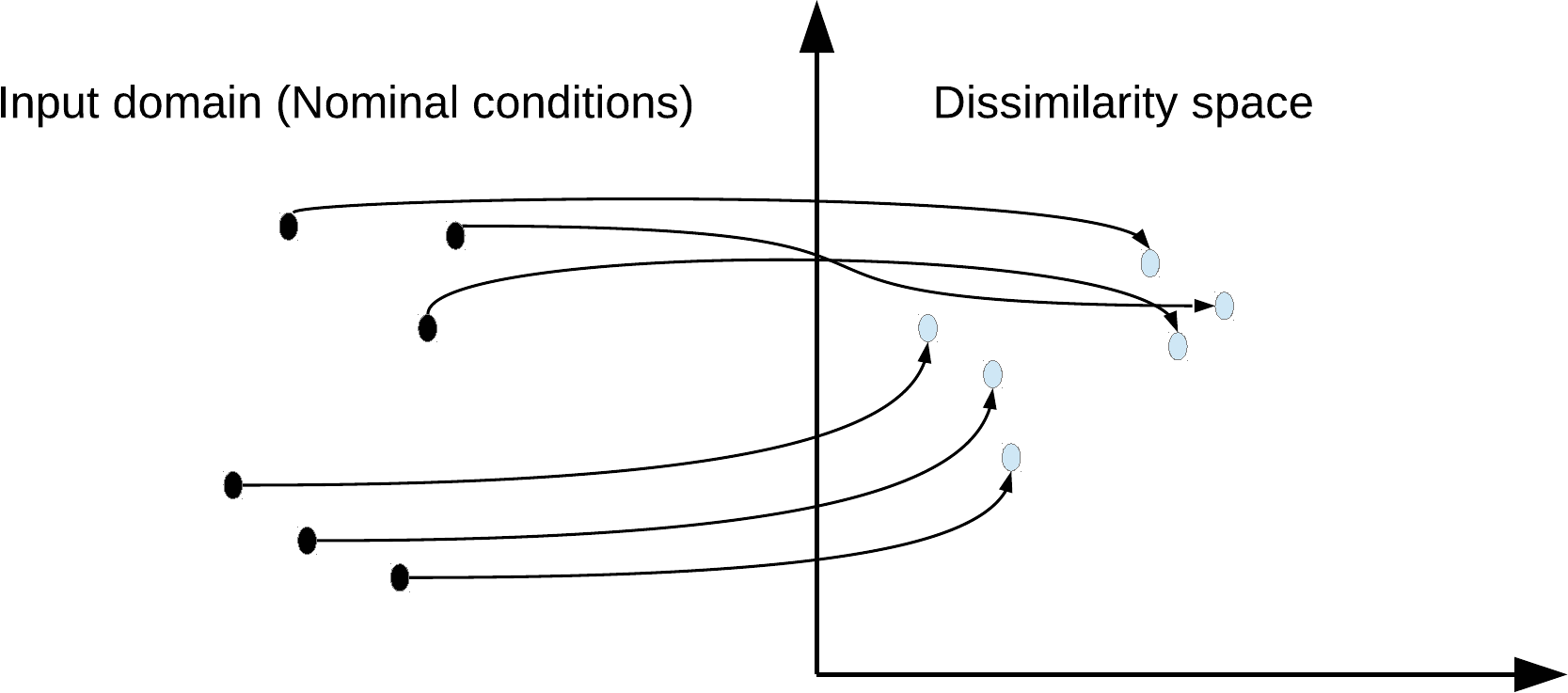}
\label{fig:f1}}

\subfigure[kNN graph construction in dissimilarity space.]{
\includegraphics[scale=0.4,keepaspectratio=true]{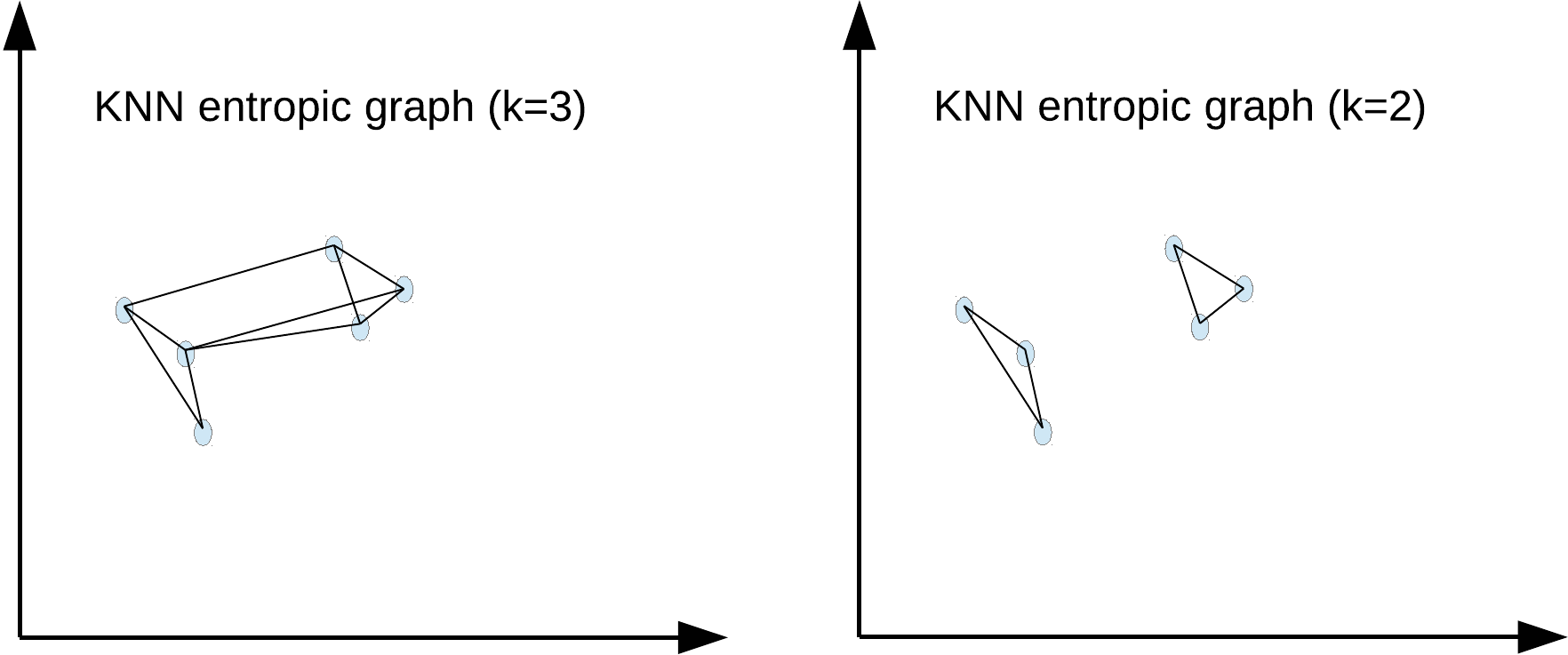}
\label{fig:f2}}

\subfigure[Graph partition and related fuzzification.]{
\includegraphics[scale=0.4,keepaspectratio=true]{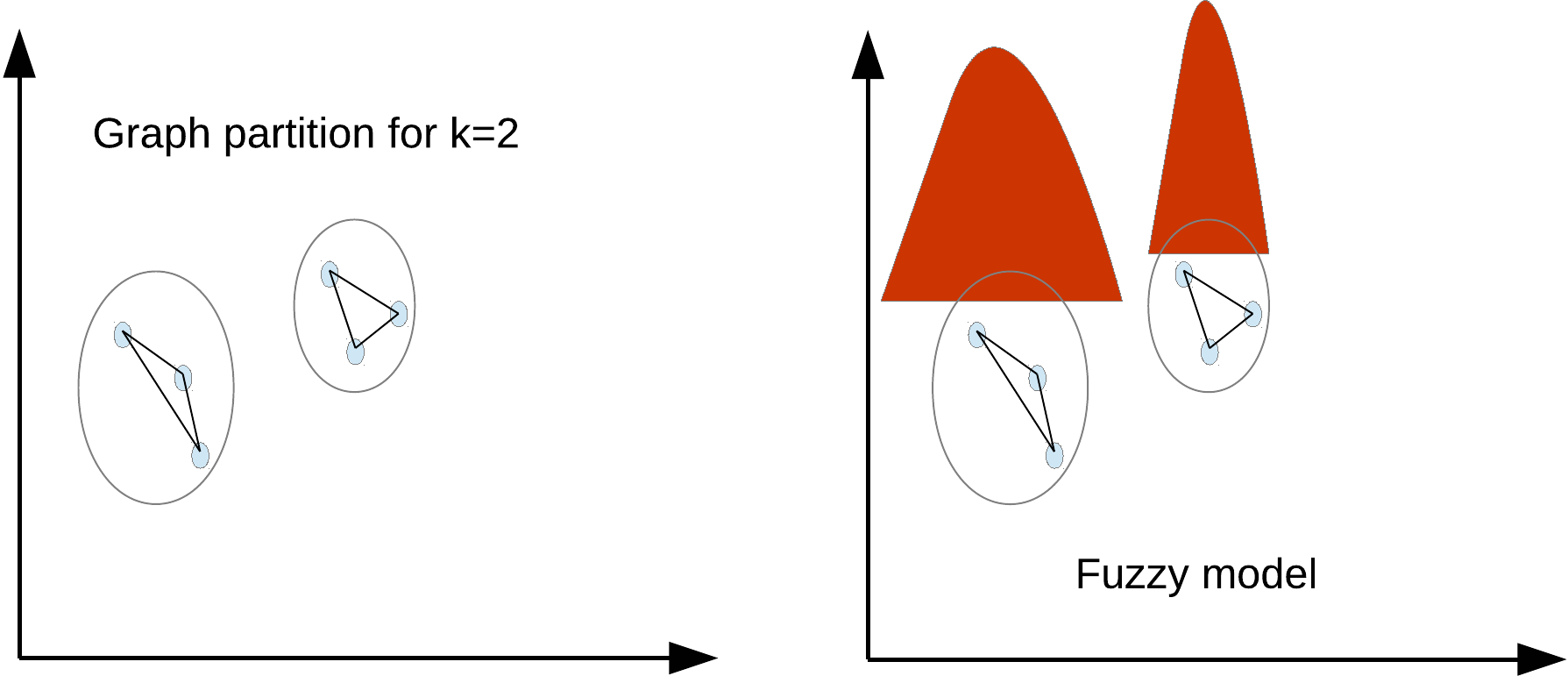}
\label{fig:f3}}

\caption{(colors online) Schematics illustrating the proposed classifier. The first step (Fig. \ref{fig:f1}) consists in mapping the input data to a dissimilarity space. A graph-based representation of the embedded data is then computed by minimizing the mutual information (Fig. \ref{fig:f2}). Finally, in order to provide a confidence level associated with the classifier decision, the embedded data are fuzzified by exploiting only topological information derived from the graph-based representation (Fig. \ref{fig:f3}).}
\label{fig:example}
\end{figure}

In this paper, we present a novel methodology for constructing one-class classifiers based on entropic spanning graphs \cite{eocc} that extends a preliminary version appeared in \cite{eocc_mi_ijcnn}.
The high-level steps behind the proposed classifier are given in Fig. \ref{fig:example}.
The first step consists in embedding the training set, representative of the nominal condition class, in an Euclidean space.
The embedding is implemented by means of the dissimilarity space representation \cite{odse} that depends on a parametric dissimilarity measure defined in the input domain.
An important consequence of this choice is that, in principle, we are able to process any input data type (e.g., graphs, sequences, and other forms of non-numerical data).
Once the embedding vectors are generated, we construct a geometric (Euclidean) graph whose vertices represent the embedded samples and the edges represent their Euclidean distances in the embedding space.
Here we use a k-nearest neighbour (kNN) graph, where $k$ is considered to be a structural parameter.
The model of the classifier is defined as one of the possible partitions of such vertices.
In particular, we derive the partition by following a criterion based on mutual information minimization.
The resulting connected components (clusters of vertices) form the decision regions (model) of the classifier.

As said before, the first step of the procedure consists in learning a dissimilarity representation of the input data.
Dissimilarity (and kernel function) learning is a well-known research field in pattern recognition \cite{schleif2015indefinite}.
In general, research efforts in the field of (dis)similarity learning focus on adapting well-known methods in the case of non-metric distances; as an alternative, embeddings are developed \cite{xiao2010geometric,wilson2014spherical} bringing back the problem to the Euclidean geometric setting.

Dissimilarity representations might yield high-dimensional embedding vectors.
Graph-based models of data are popular in pattern recognition, as in fact a graph allows to deal with the high dimensionality of the data by relying on topological information only \cite{chen2009fast,bertini2011nonparametric}.
Among the many approaches, those based on kNN graphs are particularly interesting and more related to our work.
For instance, in \cite{tomasev2014role,tomavsev2014hubness} the authors exploit the concept of ``hubness'' of samples mapped to a kNN graph for designing clustering and classification systems for high-dimensional data.

The methodology proposed here possesses several connections with other methods based on concepts borrowed from information theory \cite{chen2013system}.
In fact, information-theoretic based techniques are widely used in feature selection and image processing problems \cite{kybic2012approximate,vergara2014review,liu2009feature,sabuncu2008using,bardera2010image,brown2012conditional,plasberg2009feature,balagani2010feature}, as well as for devising clustering algorithms in both vector \cite{Vikjord20143070,alush2015pairwise,kraskov2005hierarchical,6822602,982897} and graph \cite{rosvall2007information,PhysRevE.71.046117,raj2010information,hcmst} domains.
Finally, we comment that our contribution has some affinity with the information bottleneck method \cite{still2014information,slonim2006multivariate}.
In fact, such a method prescribes that the mutual information between the input and the compressed representation should be minimized, while at the same time enforcing the maximization of the mutual information of the compressed representation with the target/output signal. Here, we exploit only the unsupervised part of this approach by computing a compressed, cluster and graph-based representation of the input.

The novelty of our contribution can be summarized as follows:
\begin{itemize}
 \item A methodology for designing one-class classifiers, where the model of the classifier is obtained by exploiting a criterion based on mutual information minimization. To this end, we use a convenient formulation for the mutual information derived from the $\alpha$-Jensen difference \cite{neemuchwala2005image}. Such a measure can be directly computed by means of a kNN-based $\alpha$-order R\'{e}nyi entropy estimator \cite{gs:HeroEtAl2002}. The final classifier, optimal according to the proposed criterion, is the one associated with the partition among all possible partitions characterised by minimum statistical dependence between the clusters.
 \item A fuzzy model built on top of the obtained graph partition that provides a way to assign a confidence level, expressed in terms of membership degree to the nominal conditions class, to each test sample.
 The fuzzification mechanism is based only on topological properties of the vertices of the entropic spanning graph. Therefore, in principle, the method is capable to model clusters with arbitrary shapes.
\end{itemize}

We show experimental results on both synthetic and real-world datasets for one-class classification, containing samples represented as feature vectors and labeled graphs.
In this paper, in addition to evaluating the method on well-known benchmarks, we also face the challenging problem of protein solubility recognition \cite{ecoli_graph_complexity}.
Classification of proteins with respect to their solubility degree is a hard yet very important scientific problem, with consequences related to the folding of such macro-molecules \cite{dill2012protein}. Here, we tackle the problem by representing a dataset of proteins in several different ways: as sequences of symbols (amino acid identifiers), as labeled graphs (hence taking into account the folded structure), as sequences of numeric vectors encoding features of such graphs, and as feature vectors extracted from the graph structures (considering also post-processed versions of such features).
The possibility to deal with a given classification problem from several different angles (i.e., by using different data representations) without the need to fine tune the method is a valuable design asset of the proposed one-class classifier.

The method presented here differs from our previous contribution \cite{eocc} in a number of ways.
First of all, here we use a kNN graph for representing the samples in the embedding space, whereas in \cite{eocc} we used a minimum spanning tree.
In addition, the objective function for learning the model is based on mutual information, while in \cite{eocc} we adopted a combination of entropy and modularity (to be maximized).
Finally, the fuzzification of the graph-based model is implemented here by using only topological information of the entropic spanning graph vertices.
In \cite{eocc}, instead, we used a more conventional centroid-based representation for the decision regions.

The remainder of this paper is structured as follows.
In Section \ref{sec:alpha_re}, we provide the technical details related to the graph-based $\alpha$-order R\'{e}nyi entropy estimator and related $\alpha$-Jensen difference. The reader might skip this section if already familiar with the basics.
Section \ref{sec:eocc_mi} provides the details on the proposed one-class classifier.
In Section \ref{sec:experiments} we discuss experimental results.
Notably, in Section \ref{sec:uci_iam_datasets} we present results on well-known benchmarking datasets (UCI and IAM).
In Section \ref{sec:protein_sol} we discuss results obtained on the problem of classifying proteins with respect to their solubility degree.
Finally, in Section \ref{sec:conclusions} we draw our conclusions and offer future research directions.

\section{R\'{e}nyi entropy graph-based estimation}
\label{sec:alpha_re}

\subsection{Graph-based estimation of R\'{e}nyi entropy}

Let $X$ be a continuous random variable with probability density function (PDF) $f(\cdot)$.
The R\'{e}nyi entropy of order $\alpha$ is defined as
\begin{equation}
\label{eq:differential_entropy}
H_{\alpha}(X)=\frac{1}{1-\alpha}\log\left(\int f(x)^{\alpha}dx\right), \ \alpha\geq0, \alpha\neq1.
\end{equation}
When $\alpha\rightarrow 1$, Eq. \ref{eq:differential_entropy} corresponds to the Shannon entropy.

Let us consider a dataset $\mathcal{S}\subset\mathbb{R}^{m}$ composed of $n$ i.i.d. $m$-dimensional realizations $\mathbf{x}_i\in \mathcal{S}, i=1, 2, ..., n$, with $m\geq 2$.
Let $G$ be an Euclidean graph, whose vertices denote the samples of $\mathcal{S}$.
An edge $e_{ij}$ connecting $\mathbf{x}_i$ and $\mathbf{x}_j$ is weighted using the Euclidean distance, $|e_{ij}| = d_{2}(\mathbf{x}_i, \mathbf{x}_j)$.
The $\alpha$-order R\'{e}nyi entropy (\ref{eq:differential_entropy}) can be estimated according to a geometric interpretation of an entropic spanning graph of $G$ \cite{Hero_Asympt__1999}.
Examples of spanning graphs used in the literature include the minimum spanning tree, kNN graph, Steiner tree, and TSP graph \cite{md_ent,Bonev2013214,pal_renyi_e_knn__2010,4897236,intrdim_shapes_hero,1326713,5967818,costa2004geodesic}.
As we will discuss more formally in the next section, entropic spanning graphs are the key elements of our contribution, which will be used also to define the model of the classifier.
Let $L_{\gamma}(G) = \sum_{e_{ij}\in G} |e_{ij}|^{\gamma}$ be the length of the entropic spanning graph defined as the sum of all weights, where $\gamma\in(0, m)$ is a user-defined parameter defining the order of the R\'{e}nyi entropy, $\alpha=(m-\gamma)/m$.
The $\gamma$ parameter allows to focus on specific weights. For instance, with $\gamma\gg 1$ it is possible to get rid of small weights in the summation; conversely, small values of $\gamma$ magnify the contribution of small weights. Although there is no optimal setting for all problems, $\gamma$ is typically set in order to obtain $\alpha=0.5$.
The R\'{e}nyi entropy of order $\alpha\in(0, 1)$ defined in (\ref{eq:differential_entropy}) can be estimated as
\begin{equation}
\label{eq:rentropy}
\hat{H}_{\alpha}(G) = \frac{m}{\gamma}\left[ \log\left(\frac{L_{\gamma}(G)}{n^{\alpha}}\right) - \log\left(\beta(L_{\gamma}(G), m)\right) \right],
\end{equation}
where $\beta(L_{\gamma}(G), m)$ is a constant term (i.e., it does not depend on the PDF) that is defined as $\beta(L_{\gamma}(G), m) \simeq \gamma/2\log\left( m/2\pi e \right)$.
The entropy estimator (\ref{eq:rentropy}) is a suitable choice for estimating information-theoretic quantities when processing high-dimensional data \cite{gs:HeroEtAl2002}.
This comment derives from the fact we use of entropic spanning graphs in (\ref{eq:rentropy}).

\subsection{R\'{e}nyi $\alpha$-divergence, $\alpha$-mutual information, and $\alpha$-Jensen difference}
\label{sec:alpha_jensen}

Starting from the $\alpha$-order R\'{e}nyi entropy (\ref{eq:differential_entropy}) it is possible to define other information-theoretic quantities, such as divergence and mutual information.
Let $f(\cdot)$ and $q(\cdot)$ be two PDFs supported on the same domain.
The R\'{e}nyi $\alpha$-divergence is defined as:
\begin{equation}
\label{eq:alpha_divergence}
D_\alpha(f\parallel q) = \frac{1}{\alpha-1} \log \int f(x)^{\alpha} q(x)^{1-\alpha} dx,
\end{equation}
with $\alpha\in(0, 1)$. Eq. \ref{eq:alpha_divergence} operates as a measure of (non metric) dissimilarity between two distributions. In fact, it is non-negative and is zero if and only if $f(\cdot)=q(\cdot)$.

The mutual information between two distributions can be used to assess their statistical dependence.
Said in other terms, mutual information can be seen as a measure of similarity between random variables, where the similarity is measured in terms of their dependence.
In fact, we remind that two distributions are statistically independent if and only if their mutual information is zero.
Let us consider our dataset $\mathcal{S}$.
Calculating the $\alpha$-divergence between the joint and product of marginal distributions of $\mathcal{S}$ features allows to define the so-called $\alpha$-mutual information \cite{neemuchwala2005image}.
Such a measure computes the degree of statistical dependence between the components forming $\mathcal{S}$ (or, more generally, between $d$ different random variables defined on the same domain).
This is obtained by assessing $f(\cdot)$ and $q(\cdot)$ in (\ref{eq:alpha_divergence}) as the joint distribution and product of the marginals of $\mathcal{S}$, respectively.
Hence, mutual information and divergence provide powerful and complimentary tools to define (dis)similarity based recognition systems (in our case, one-class classifiers).

The $\alpha$-Jensen difference, with $\alpha\in(0, 1)$, is a useful measure of dissimilarity between two distributions that is defined as
\begin{equation}
\label{eq:alpha_jensen}
\begin{aligned}
&\Delta H_{\alpha}(\beta, f, q) = \\
&H_{\alpha}(\beta f + (1-\beta)q) - [\beta H_{\alpha}(f) + (1-\beta)H_{\alpha}(q) ],
\end{aligned}
\end{equation}
where $\beta\in[0, 1]$ allows for a linear convex combination of the PDFs.
Since the $\alpha$-order R\'{e}nyi entropy (\ref{eq:differential_entropy}) is strictly concave in $f(\cdot)$ for $\alpha\in(0, 1)$ \cite{neemuchwala2005image}, the well-known Jensen's inequality assures that Eq. \ref{eq:alpha_jensen} is non-negative and degenerates to zero if and only if $f(\cdot)=q(\cdot)$.
This fact suggests to consider the $\alpha$-Jensen difference (\ref{eq:alpha_jensen}) as a reliable alternative for the $\alpha$-divergence (\ref{eq:alpha_divergence}) and hence also as a measure of statistical dependence between distributions.
In particular, Eq. \ref{eq:alpha_jensen} is defined by considering only (combinations of) $\alpha$-order R\'{e}nyi entropy.
Therefore, the $\alpha$-Jensen difference can be estimated directly by using bypass estimators for each entropy term appearing in Eq. \ref{eq:alpha_jensen}, such as the graph-based entropy estimator shown in Eq. \ref{eq:rentropy}.
Moreover, by Jensen inequality, it can be easily extended in order to handle $d$ different PDFs as follows,
\begin{equation}
\label{eq:alpha_jensen_graph}
\Delta \hat{H}_{\alpha}(\beta, G, d) = \hat{H}_{\alpha}\left(G\right) - \left[\sum_{i=1}^d \beta_i \hat{H}_{\alpha}(G_i) \right],
\end{equation}
where $\beta_i=|G_i|/|G|$ and $|G_i|$ indicates the number of vertices in $G_i$, guaranteeing thus $\sum_{i=1}^{d} \beta_i = 1$.
In Eq. \ref{eq:alpha_jensen_graph}, $G_i, i=1, ..., d$, are $d$ sub-graphs of $G$ representing the $d$ different PDFs under considerations.

Here we use Eq. \ref{eq:alpha_jensen_graph} as a measure of (statistical) dissimilarity between the $d$ sub-graphs $G_i$ extracted from $G$ derived during the synthesis phase -- see the following section.

\section{A one-class classifier based on mutual information minimization}
\label{sec:eocc_mi}

The section describes our contribution.
First, in Sec. \ref{sec:highlevel_description} we provide a high-level description of the main steps characterizing the proposed method.
In Sec. \ref{sec:min_mi} and \ref{sec:fuzzy_model} we describe, respectively, the synthesis of the model, the graph-based fuzzification and how to use the classifier in the operational test modality.
Finally, in Sec. \ref{sec:comp_complex} we analyze the asymptotic computational complexity.

\subsection{High-level description of the proposed method}
\label{sec:highlevel_description}

\begin{figure}[ht!]
 \centering
 \includegraphics[scale=0.39,keepaspectratio=true]{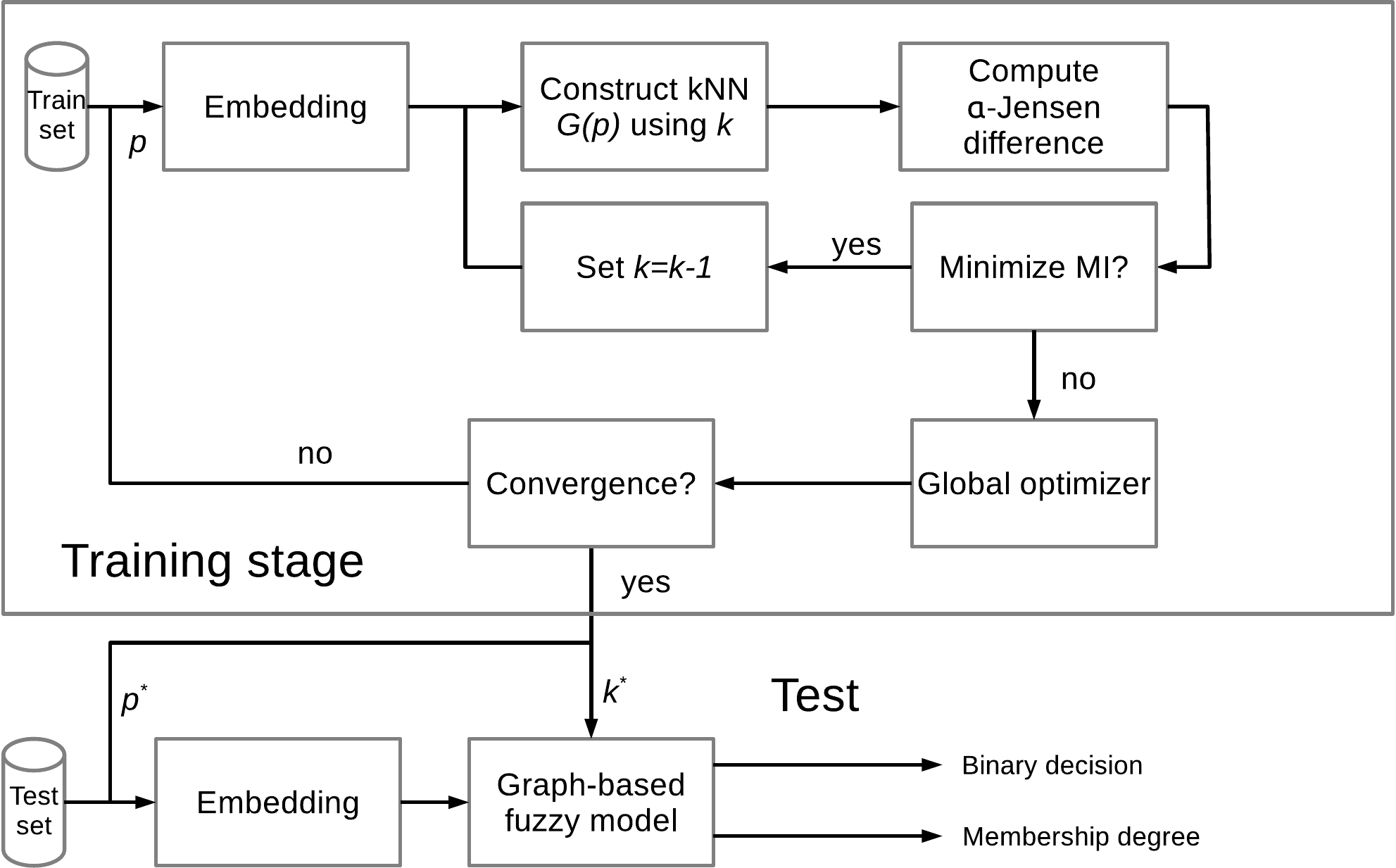}
 \caption{Block scheme describing training and testing phases of the proposed one-class classifier.}
 \label{fig:bloch_scheme}
\end{figure}

Fig. \ref{fig:bloch_scheme} provides a block scheme describing the main steps of the proposed one-class classifier.
The classifier is applicable to any input domain $\mathcal{X}$; as such, it operates also in domains of non-geometric data (e.g., non-numeric data such as labeled graphs).
Such a goal is achieved by first embedding the input dataset $\mathcal{S}\subset\mathcal{X}, |\mathcal{S}|=n$, into a $m$-dimensional Euclidean space.
The embedding step can be described by a map $\phi: \mathcal{P}(\mathcal{X})\times\mathcal{P}(\mathcal{X})\times\Gamma\rightarrow\mathbb{R}^{n\times m}$, where $\mathcal{P}(\mathcal{X})$ is the power-set of $\mathcal{X}$ and $\Gamma$ is a domain of numeric parameters of the input dissimilarity measure.
A set of prototypes, $\mathcal{R}\subseteq\mathcal{S}, |\mathcal{R}|=m\leq n$, called representation set, is used to compute the dissimilarity matrix, $\mathbf{D}^{n\times m}$, given as $D_{ij}=d_{\mathrm{I}}(x_i, r_j)$, $x_i\in\mathcal{S}$ and $r_j\in\mathcal{R}$, where $d_{\mathrm{I}}: \mathcal{X}\times\mathcal{X}\rightarrow\mathbb{R}^{+}$ is a non-negative (bounded) dissimilarity measure.
In order to give more flexibility to the method, we assume that $d_{\mathrm{I}}(\cdot, \cdot)$ depends on some (numerical) parameters, say $p\in\Gamma$ which, in turn, influence the resulting vector configuration in the embedding space.
Developing the dissimilarity space representation (DSR) of $\mathcal{S}$ is a straightforward yet principled way to construct an Euclidean embedding space.
Each input sample $x_i\in\mathcal{S}$ is represented by the corresponding row-vector $\mathbf{x}_i$ of matrix $\mathbf{D}$.
Therefore, the embedding step for the entire $\mathcal{S}$ is formally described as $\mathbf{D}=\phi(\mathcal{S}, \mathcal{R}, p)$.
This stresses that the embedding is controlled/influenced by both $\mathcal{R}$ and $p$.
An important property of the DSR is that, if $d_{\mathrm{I}}(\cdot, \cdot)$ is metric, distances in the dissimilarity space are preserved up to a scaling factor equal to $\sqrt{m}$, that is, $d_2(\mathbf{x}_i, \mathbf{x}_j)\leq d_{\mathrm{I}}(x_i, x_j) \sqrt{m}$ (Lipschitz mapping).
On the other hand, if $d_{\mathrm{I}}(\cdot, \cdot)$ is not metric (for instance it violates the triangular inequality), it is possible to prove that such a property can be still satisfied by considering a constant related to the maximum violation of the triangular inequality \cite{pkekalska+duin2005}.

Embedded data are then represented by means of a geometric graph, $G=(\mathcal{V}, \mathcal{E})$.
In this paper, we rely on kNN graphs.
In kNN graphs, vertices $\mathcal{V}$ denote embedded samples in $\mathcal{S}$, while edges their relations in terms of distance; each edge $e_{ij}\in\mathcal{E}$ is assigned a weight $w_{ij}=d_2(\mathbf{x}_i, \mathbf{x}_j)$.
The dimensionality of an DSR depends on the number of prototypes used during the computation of the dissimilarity matrix (the number of prototypes is the number of dimensions).
Determination of such prototypes is an important yet complex task; it is equivalent to a problem of feature selection.
Several methods have been proposed in the literature (see \cite{odse} and references therein) to address this problem. However, all such methods are non-trivial in terms of computations, especially when contextualized within a more complex system.
Therefore, our choice was to avoid such computations and consider simple initializations of the prototypes -- see in the following.
Nonetheless, constructing $G$ on the DSR allows us also to overcome/mitigate the problem of prototype selection, as in fact graphs are suitable for representing (possibly) high-dimensional data.
In the following, we denote with $G(p)$ the Euclidean graph constructed over the DSR of $\mathcal{S}$ computed using $p$ for $d_{\mathrm{I}}(\cdot, \cdot)$.

We define the model of the classifier in terms of decision regions by using the structural information derived from the kNN graph $G(p)$.
In particular, each connected component of $G(p)$ forms a decision region.
We propose a synthesis procedure for learning such decision regions based on optimization of the $\alpha$-Jensen difference (\ref{eq:alpha_jensen_graph}).
In doing so, we find a partition of the graph whose components denote high divergence, i.e., high statistical dissimilarity.

In order to provide a confidence level associated with the classifier decision, once training is complete, we fuzzify the obtained graph-based representation $G(p)$. In practice, a membership degree is assigned to each vertex of the kNN graph. Such membership degrees are based only on the topological information provided by $G(p)$, hence allowing to model data with complex geometric structures.

A test sample is classified during the operational modality by first mapping it into the DSR. Then, the sample is fuzzified by using the related topological information. The classifier provides two types of decisions regarding the membership of test samples to the nominal conditions class: a binary and a membership degree.

\subsection{Training by minimizing the mutual information}
\label{sec:min_mi}

We propose a learning approach for synthesizing the classifier model by exploiting the $\alpha$-Jensen difference (\ref{eq:alpha_jensen_graph}) using a kNN graph.
Let us define the objective function
\begin{equation}
\label{eq:objective_function}
\eta(p, k)=\frac{1}{1+\Delta \hat{H}_{\alpha}(\beta, G(p), d(k))} \in [0, 1],
\end{equation}
which basically converts the $\alpha$-Jensen difference (\ref{eq:alpha_jensen_graph}) in a similarity measure.
Although, formally speaking, Eq. \ref{eq:objective_function} cannot be thought as a form of $\alpha$-mutual information, we argue that it could be used as a proxy for its computation.
In fact, Eq. \ref{eq:objective_function} assumes one when the divergence is zero (i.e., when all distributions are equal) and tends to zero as the divergence grows (see \cite{neemuchwala2005image} for further examples).
The optimization problem describing the one-class classifier model synthesis is hence formulated as
\begin{equation}
\label{eq:opt_minimization_alpha_jensen}
\min_{k,p} \eta(p, k),
\end{equation}
where $k\geq1$ is the parameter of the kNN spanning graph that acts here a structural parameter, defining the upper bound for the partition order $d$.
In fact, the number $d=d(k)\geq1$ of connected components (sub-graphs) of the resulting kNN graph depends on $k$, as made explicit in (\ref{eq:objective_function}).
Parameters $p$ affect the dissimilarity measure used to construct the DSR.
For instance, when the input domain $\mathcal{X}$ corresponds to $\mathbb{R}^n$, then $d_{\mathrm{I}}(\cdot, \cdot)$ could be defined as an Euclidean metric with $p$ as a vector of weights (or as a covariance matrix).
Therefore, the choice of $p$ influences also the topology of the kNN graph.
With (\ref{eq:opt_minimization_alpha_jensen}) we propose to minimize the dependence (or equivalently, maximize the independence) between the distribution of the resulting connected components, i.e., the clusters of connected vertices derived from the kNN graph.
The optimization problem (\ref{eq:opt_minimization_alpha_jensen}) consists in searching for the model order $d$ through $k$ and the parameters $p$ of the dissimilarity measure that maximize the estimated $\alpha$-Jensen difference (\ref{eq:alpha_jensen_graph}) calculated on the resulting graph partition.
Please notice that valid solutions for (\ref{eq:opt_minimization_alpha_jensen}) include also kNN graphs with only one connected component (i.e., fully connected graphs).

Algorithm \ref{alg:mi_minimization} delivers the pseudo-code describing the proposed model synthesis strategy.
A partition $P(G(p))$ is formed by means of the connected components of the kNN graph.
A global loop at line \ref{alg:global_loop} searches for the parameters $p$ (that can be a vector) of the input dissimilarity measure affecting the DSR of $\mathcal{S}$ and hence the resulting kNN graph.
In this paper, all input samples are used as prototypes for the DSR, i.e., $\mathcal{R}=\mathcal{S}$.
The high-dimensionality of the resulting DSR does not pose a serious problem, since all operations are based on the kNN graph.
However, when the cardinality of $\mathcal{S}$ is large, we sub-sample the dataset with a randomized selection scheme.
At line \ref{alg:global_k}, we perform an additional loop where we decrease the structural parameter $k$, hence forming a non-decreasing number $d\leq k$ of connected components (sub-graphs) in the related kNN graph.
Within this loop, we use Eq. \ref{eq:alpha_jensen_graph} to estimate the $\alpha$-Jensen difference in the dissimilarity space.
The entropy of the entire dataset, $\hat{H}_{\alpha}\left(G\right)$, is estimated using $k+1$ for the kNN graph.
This is performed to allow for a different number of edges in the resulting kNN graph with respect to the total number of edges in the sub-graphs derived by using $k$.

The algorithm proceeds iteratively by increasing the structural complexity of the model, i.e., by decreasing $k$.
We use a heuristic at line~\ref{lst:line:greedy} to terminate the search: when Eq. \ref{eq:objective_function} starts to increase, then the model complexity should not be increased further.
This choice is justified by considering the nature of our problem (\ref{eq:opt_minimization_alpha_jensen}) which, in fact, consists solely in minimizing the statistical dependence.
A global convergence criterion (line \ref{alg:global_convergence}) is used to stop the search on the parameters $p$.
Since our objective is to be able to process any type of input data, here we use a derivative-free approach for searching for best-performing $p$.
However, different search methods (e.g., gradient-based approaches) could be used by considering specific problem instances, such as when processing feature vectors.
The global convergence criterion is activated when a maximum number of iterations is performed or the objective function falls below a threshold $\tau\simeq 0$.
\begin{algorithm}[ht!]\footnotesize
\caption{Training of the one-class classifier.}
\label{alg:mi_minimization}
\begin{algorithmic}[1]
\REQUIRE A dataset $\mathcal{S}$ of $n$ samples
\ENSURE $P(G(p^{*}))$
\STATE Determine the prototypes $\mathcal{R}\subseteq\mathcal{S}$
\WHILE{Global optimization cycle}\label{alg:global_loop}
\STATE Get an instance of the parameters, $p$
\STATE Construct the DSR of $\mathcal{S}$ using $\mathcal{R}$ and $p$
\FOR{$k=\sqrt{n}, ..., 1$}\label{alg:global_k}
\STATE Construct the kNN spanning graph using $k$
\STATE Derive the partition, $P(G(p))_{k}$, by grouping vertices according to the resulting connected components
\STATE Compute the $\alpha$-divergence (\ref{eq:alpha_jensen_graph})
\STATE Evaluate objective (\ref{eq:objective_function})
\IF{$\eta(p, k)>\eta(p, k+1)$} \label{lst:line:greedy}
\STATE Exit for loop at line \ref{alg:global_k}
\ENDIF
\ENDFOR
\IF{Global convergence}\label{alg:global_convergence}
\STATE Store best $k^{*}=k$ and $p^{*}=p$
\RETURN $P(G(p^{*}))$ with $d=d(k^{*})$ components
\ENDIF
\ENDWHILE
\end{algorithmic}
\end{algorithm}

\subsection{Fuzzy model based on vertex centrality}
\label{sec:fuzzy_model}

We exploit the graph-based representation $G(p^{*})$, where $p^{*}$ denotes the optimal dissimilarity measure parameters, for constructing a fuzzy model.
A fuzzy set is assigned to each of the $d$ sub-graphs, by defining the fuzzification mechanism based only on the topological importance (centrality) of vertices.
In particular, we use the vertex closeness centrality as a measure of importance of the vertices.

Let $G_i$ be the $i$th sub-graph of $G$ (note we omit parameters $p^{*}$ only for the sake of simplicity).
The closeness centrality of a vertex $v$ is computed as
\begin{equation}
\label{eq:v_closeness_centrality}
\chi_i(v) = \sum_{u\neq v} 2^{-d_{G_{i}}(v, u)},
\end{equation}
where $d_{G_{i}}(v, u)$ is the weighted topological distance between vertices $v$ and $u$ in $G_{i}$.
We remind that $G_{i}$ is a geometric graph constructed over a dissimilarity space.
Therefore each edge has a weight that is given by the Euclidean distance among the corresponding samples in the dissimilarity space.

Let $\chi_{i}^{*}=\max_{v} \chi_{i}(v)$ be the maximum closeness centrality value for the $i$th sub-graph.
We define $\hat{\chi}_{i}(v)$ and $\hat{\chi}_{i}$ respectively as the vertex-specific difference and $l$th percentile of all differences with respect to the maximum closeness centrality value, i.e.,
\begin{align}
\label{eq:CCDiff}
\hat{\chi}_{i}(v) &= \chi_{i}^{*}-\chi_i(v); \\
\label{eq:CCDiff_stats}
\hat{\chi}_{i} &= \mathrm{percentile}_{v}(\hat{\chi}_{i}(v), l).
\end{align}
Note that $l$ in (\ref{eq:CCDiff_stats}) can be adjusted depending on the problem at hand; for instance, $l=50$ gives the median value.
The membership degree of $v$ to the $i$th sub-graph is defined as:
\begin{equation}
\label{eq:membership}
\mu_i(v) = \exp\left(-\frac{\hat{\chi}_{i}(v)^{2}}{2\hat{\chi}_{i}^2}\right).
\end{equation}
It is important to note that, as a consequence of the definition of Eqs. \ref{eq:v_closeness_centrality}--\ref{eq:membership}, there might be more than one vertex in a sub-graph with membership equal to 1 (i.e., with equal closeness centrality).

When a new test sample $x$ has to be evaluated, we assign to it the maximum among the membership values computed for each of the $d$ sub-graphs:
\begin{equation}
\label{eq:maximum_membership}
\mu(x) = \max_{i=1, 2, ..., d} \mu_i(x).
\end{equation}

Eq. \ref{eq:maximum_membership} is used to define the membership of test samples to the class of nominal data: 0 implies that a sample does not belong to that class, 1 indicates that it completely belongs to it, while everything in-between provides a membership degree to the class of nominal data.

The binary decision rule, instead, operates by checking if
\begin{equation}
\label{eq:binary_decision}
\hat{\chi}_{j}(x) \leq \hat{\chi}_{j},
\end{equation}
where $j=\argmax_{i=1, 2, ..., d} \mu_i(x)$.

In order to implement Eq. \ref{eq:maximum_membership}, we need to (i) embed $x$ in the dissimilarity space induced by using $p^{*}$ and (ii) add a vertex corresponding to $x$, say $v_x$, iteratively to each sub-graph $G_i, i=1, 2, ..., d$, with $k^{*}$ new edges, where $p^{*}$ and $k^{*}$ are derived during training.
It is worth noting an important fact here. When $v_x$ is added to $G_i$, the resulting kNN graph might change globally: several nearest neighbour relations might be affected.
Therefore, once $v_x$ is added to $G_i$, we re-compute the kNN graph and perform the operations described by Eqs. \ref{eq:v_closeness_centrality}--\ref{eq:binary_decision}.

\begin{figure*}[ht!]
\centering

\subfigure[Sample dataset.]{
\includegraphics[scale=0.35,keepaspectratio=true]{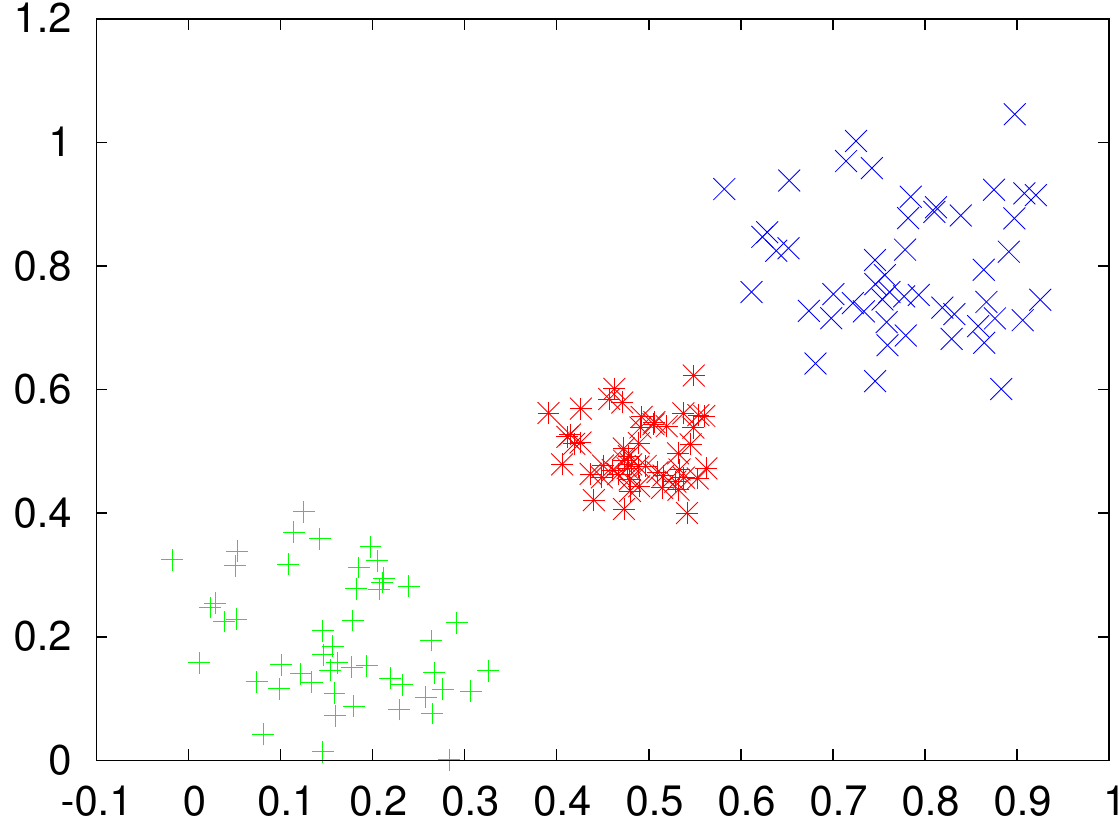}
\label{fig:ds1}}
~
\subfigure[Graph-based model.]{
\includegraphics[scale=0.35,keepaspectratio=true]{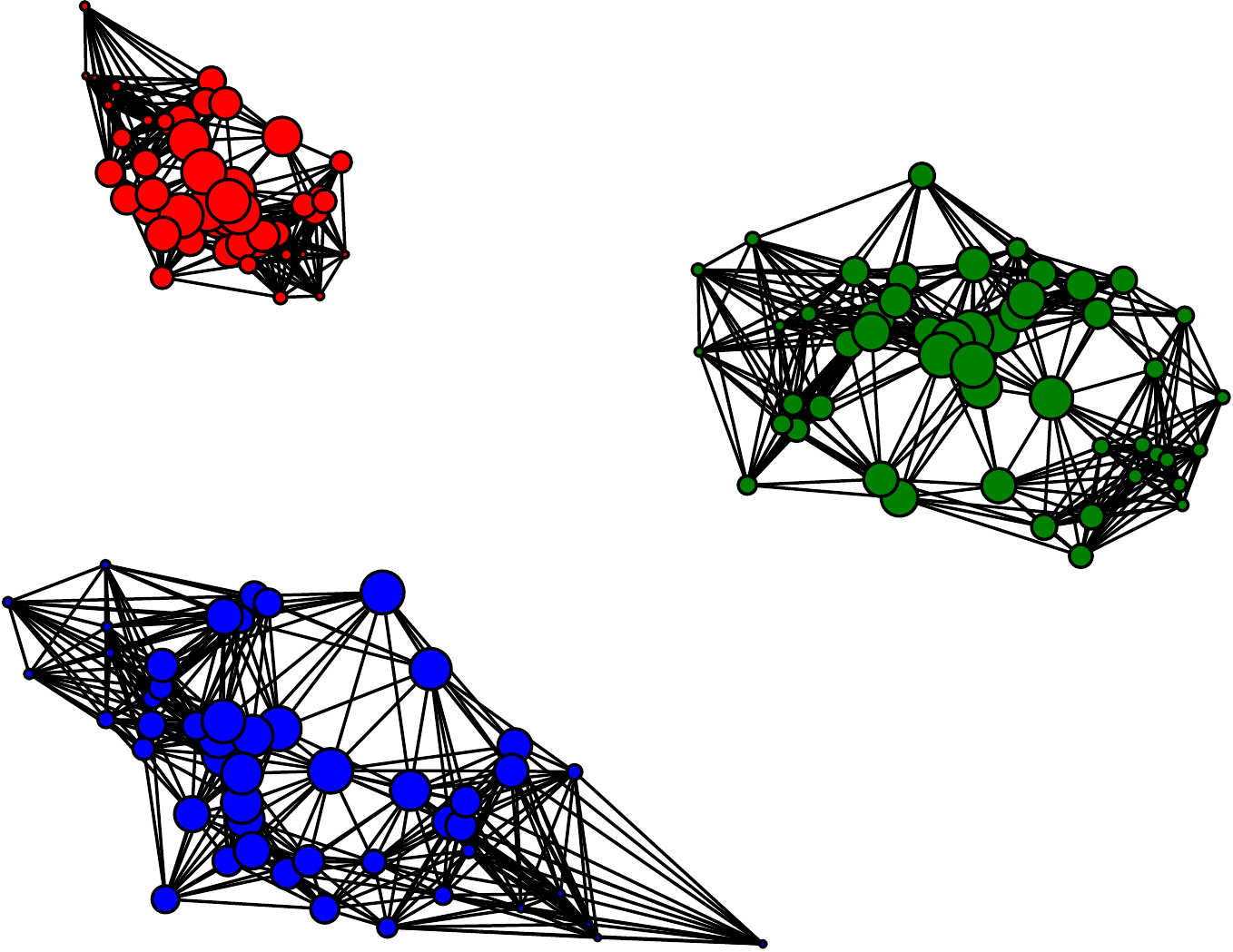}
\label{fig:ds1_mv}}
~
\subfigure[Density of differences for each sub-graph.]{
\includegraphics[width=0.32\textwidth,keepaspectratio=true]{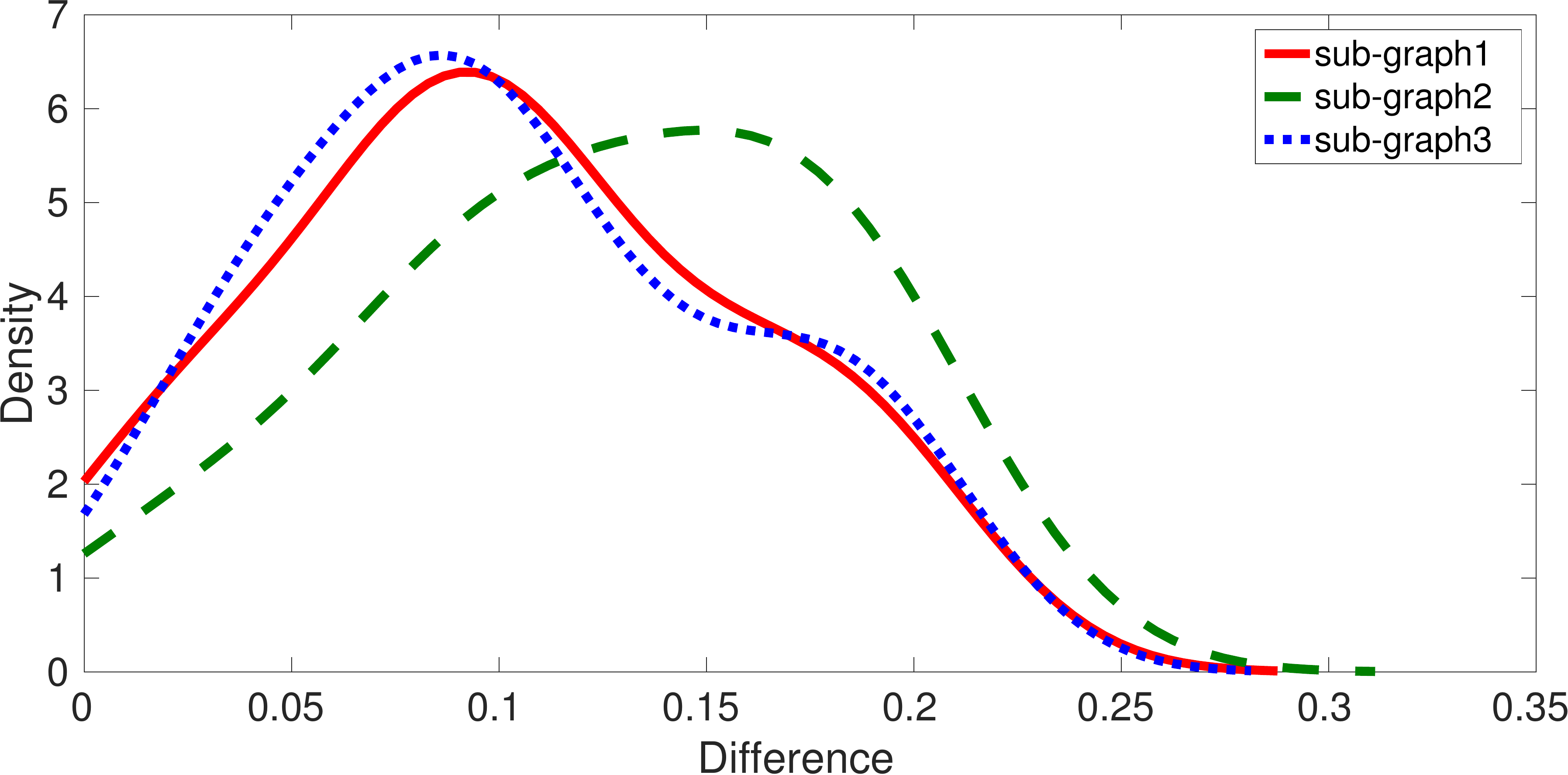}
\label{fig:ds1_CCDiff_density}}

\subfigure[Uniform dataset.]{
\includegraphics[scale=0.35,keepaspectratio=true]{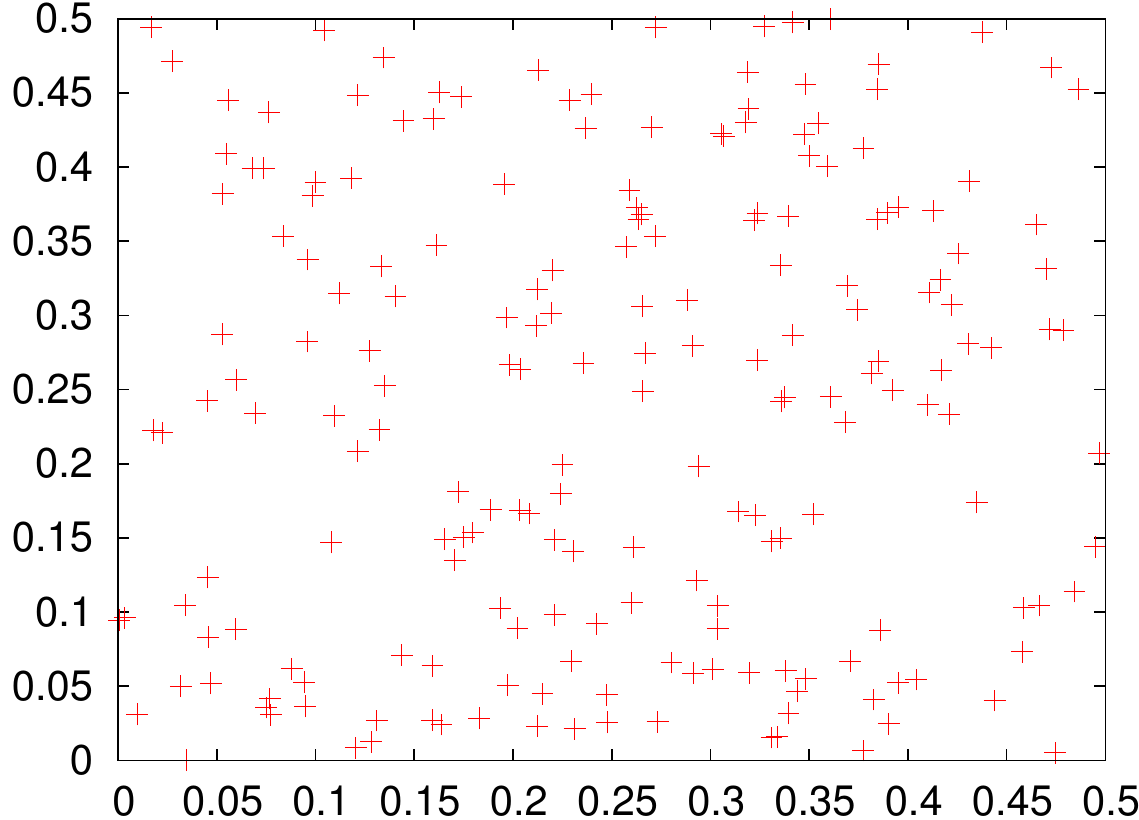}
\label{fig:ds4}}
~
\subfigure[Graph-based model.]{
\includegraphics[scale=0.35,keepaspectratio=true]{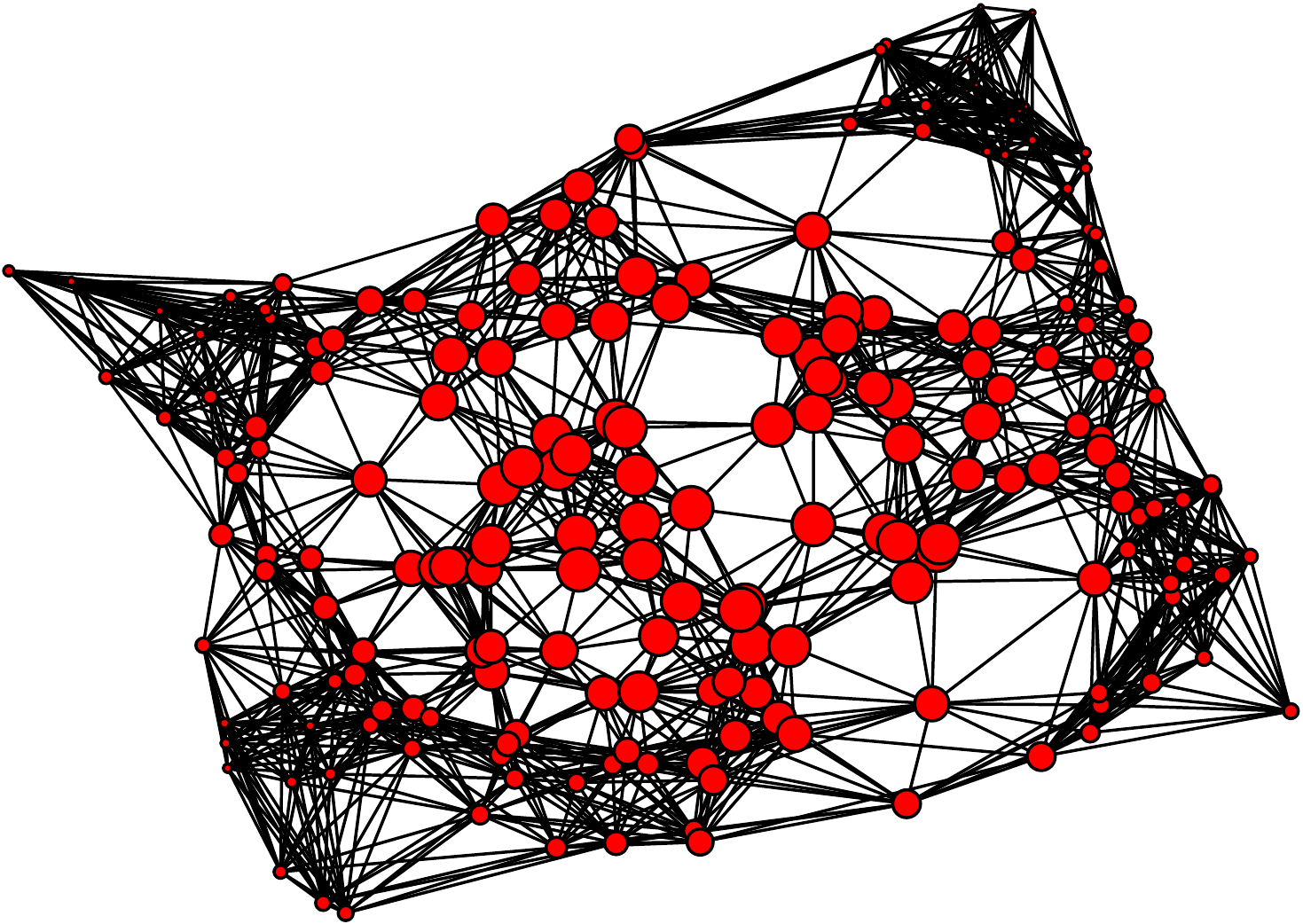}
\label{fig:ds4_mv}}
~
\subfigure[Density of differences.]{
\includegraphics[width=0.32\textwidth,keepaspectratio=true]{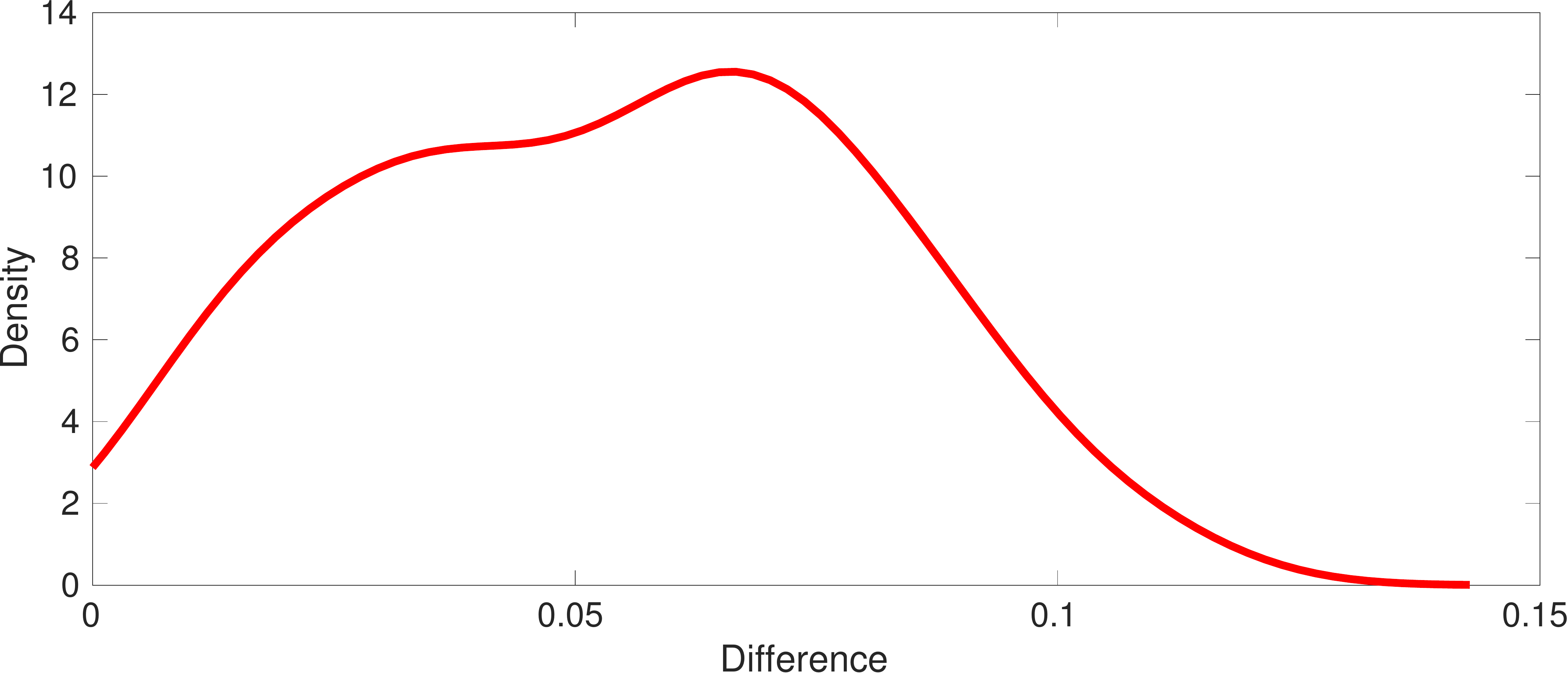}
\label{fig:ds4_CCDiff_density}}

\subfigure[Crescent-full moon dataset.]{
\includegraphics[scale=0.35,keepaspectratio=true]{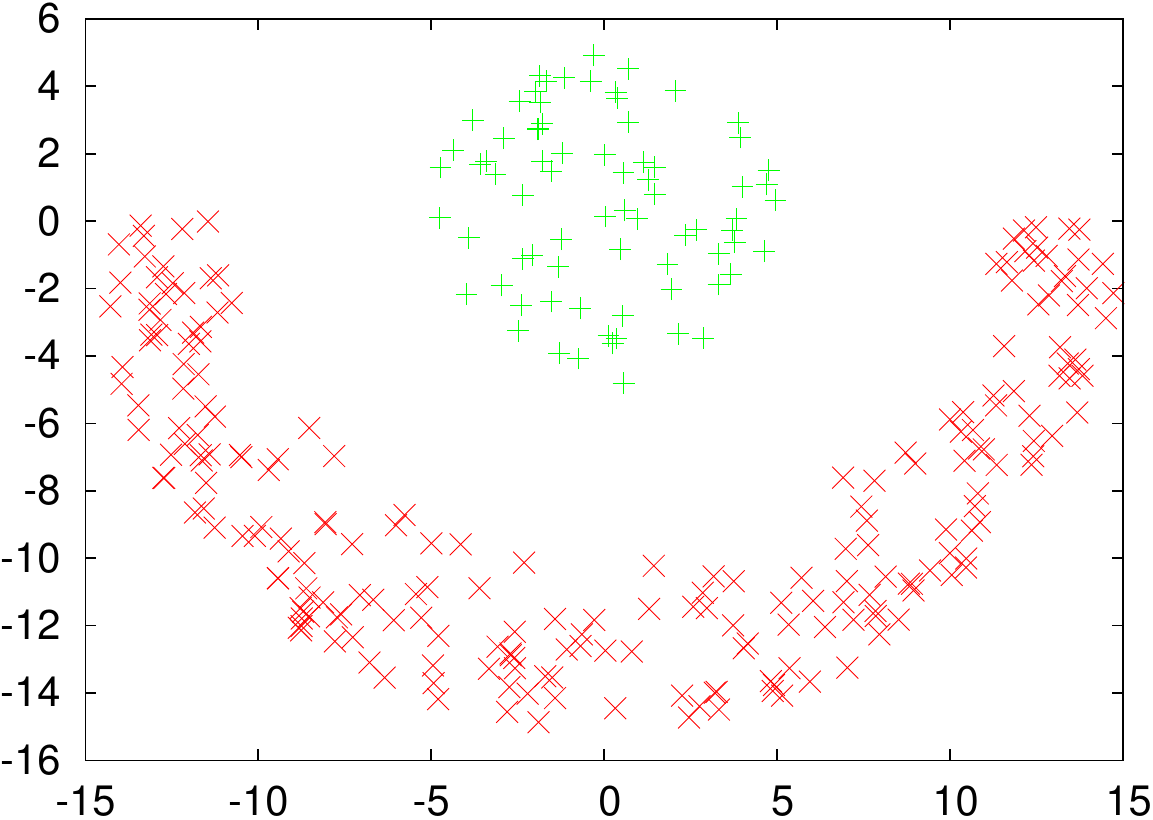}
\label{fig:ds_crescentfullmoon}}
~
\subfigure[Graph-based model.]{
\includegraphics[scale=0.35,keepaspectratio=true]{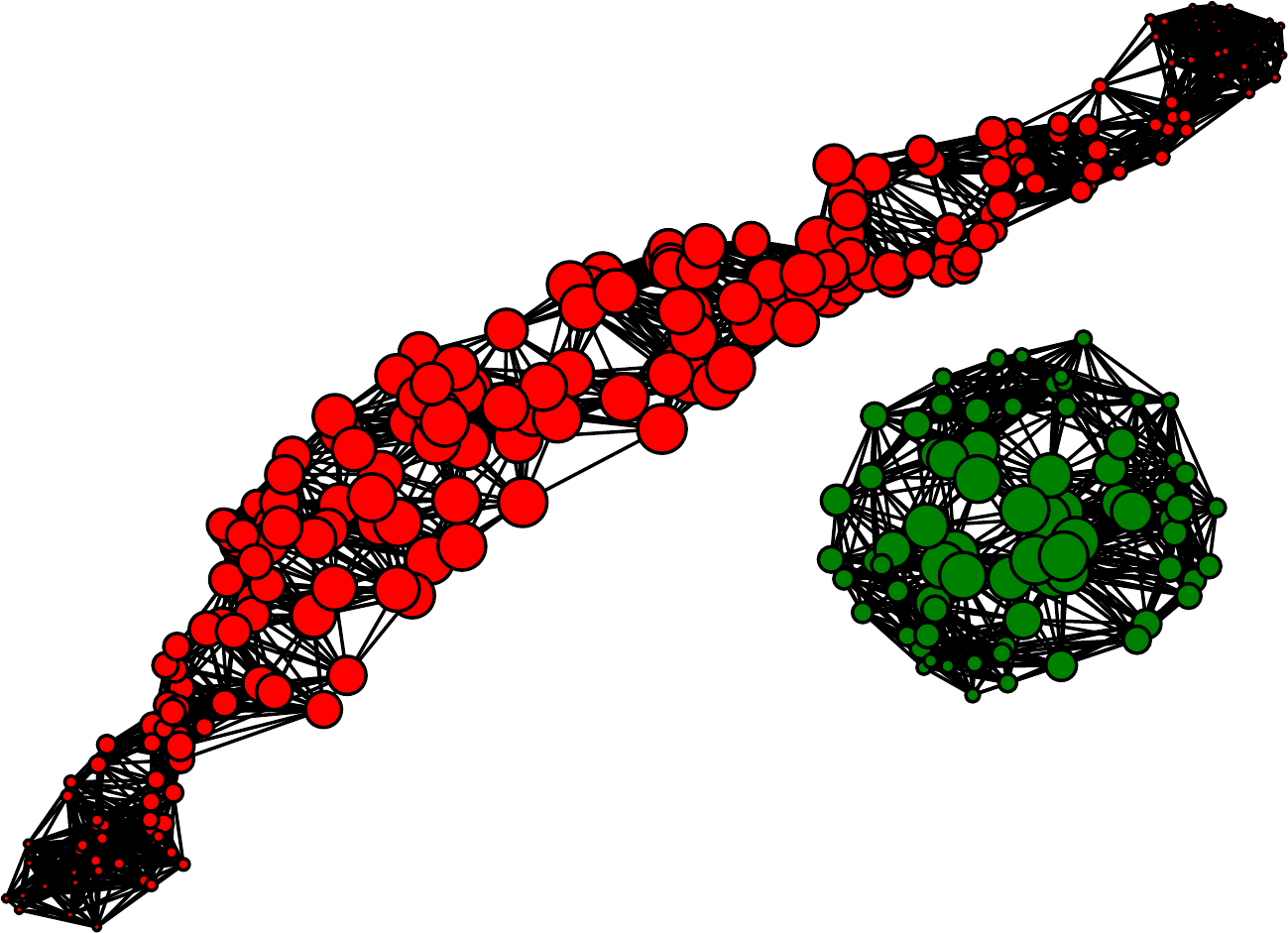}
\label{fig:ds_crescentmoon_mv}}
~
\subfigure[Density of differences for each sub-graph.]{
\includegraphics[width=0.33\textwidth,keepaspectratio=true]{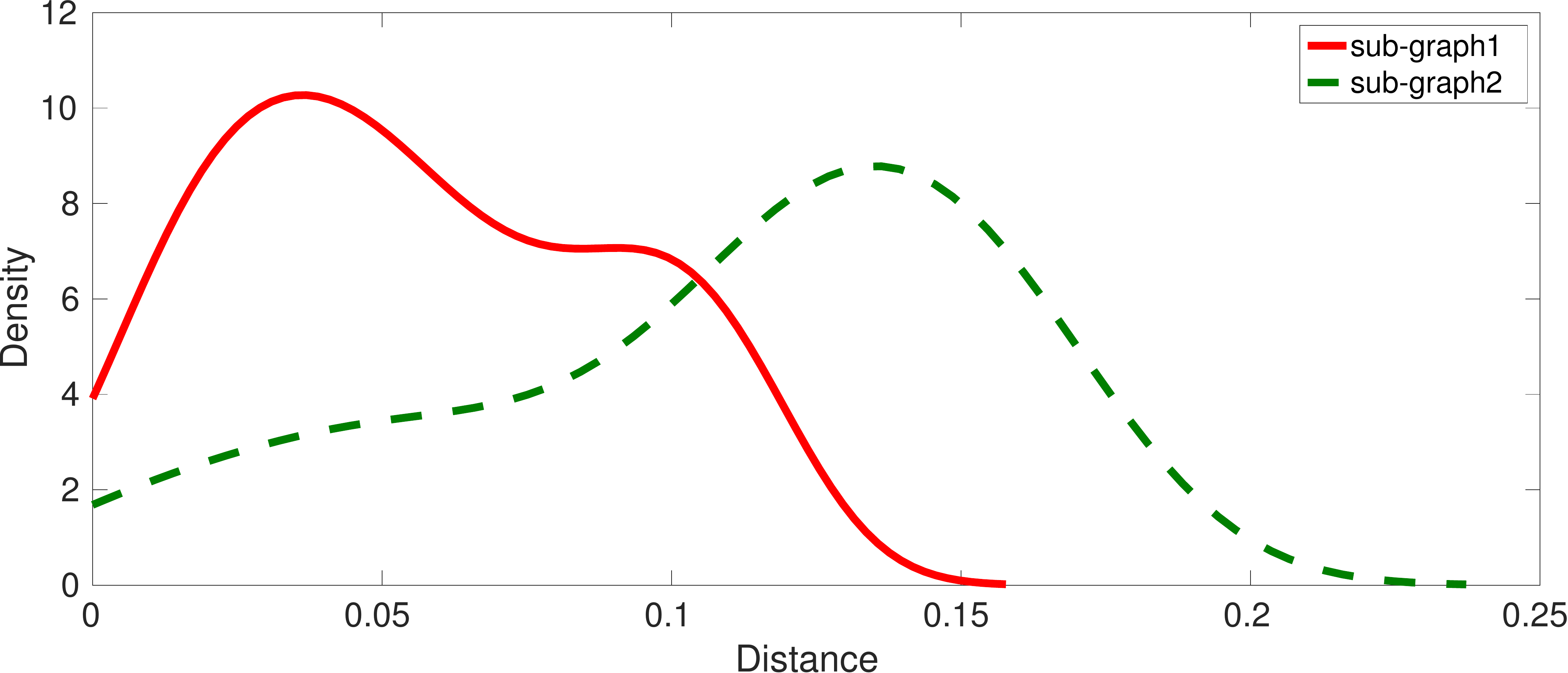}
\label{fig:ds_crescentmoon_CCDiff_density}}

\subfigure[First two principal components of a 100-dimensional dataset.]{
\includegraphics[scale=0.35,keepaspectratio=true]{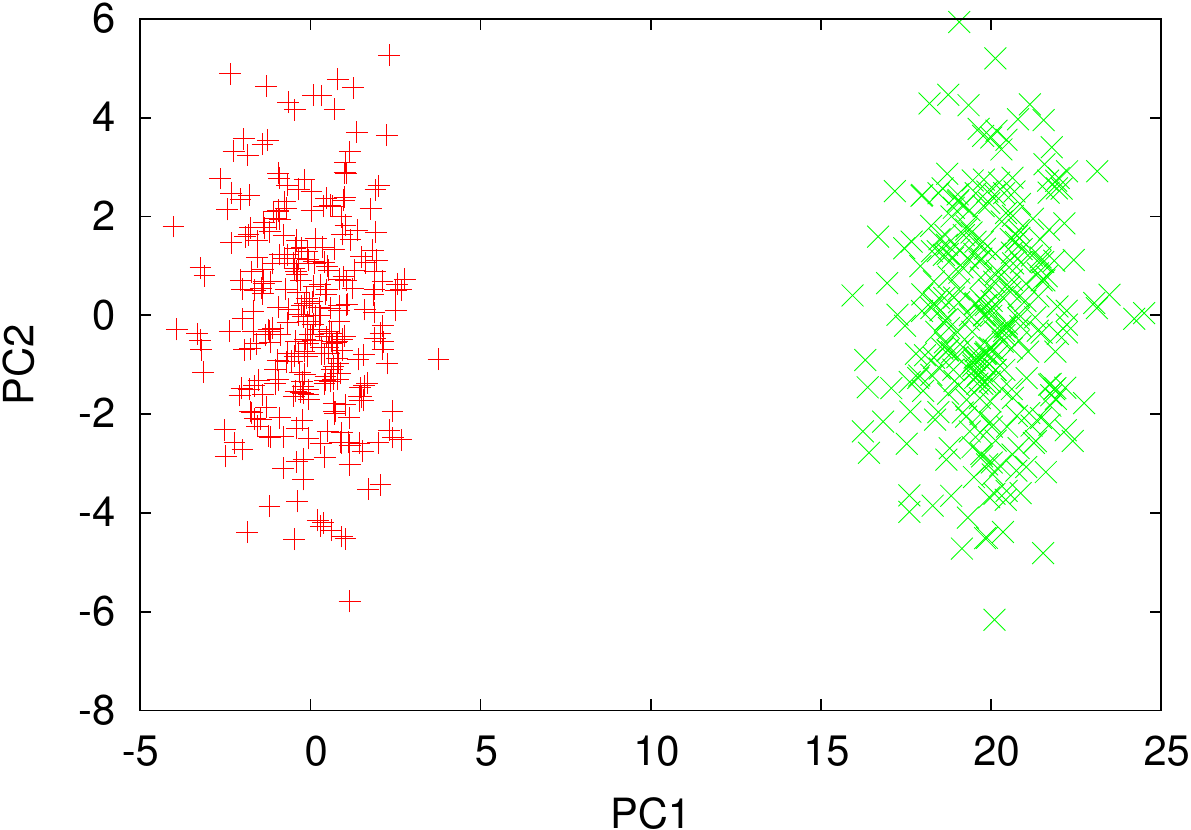}
\label{fig:ds100dim_pca}}
~
\subfigure[Graph-based model.]{
\includegraphics[scale=0.35,keepaspectratio=true]{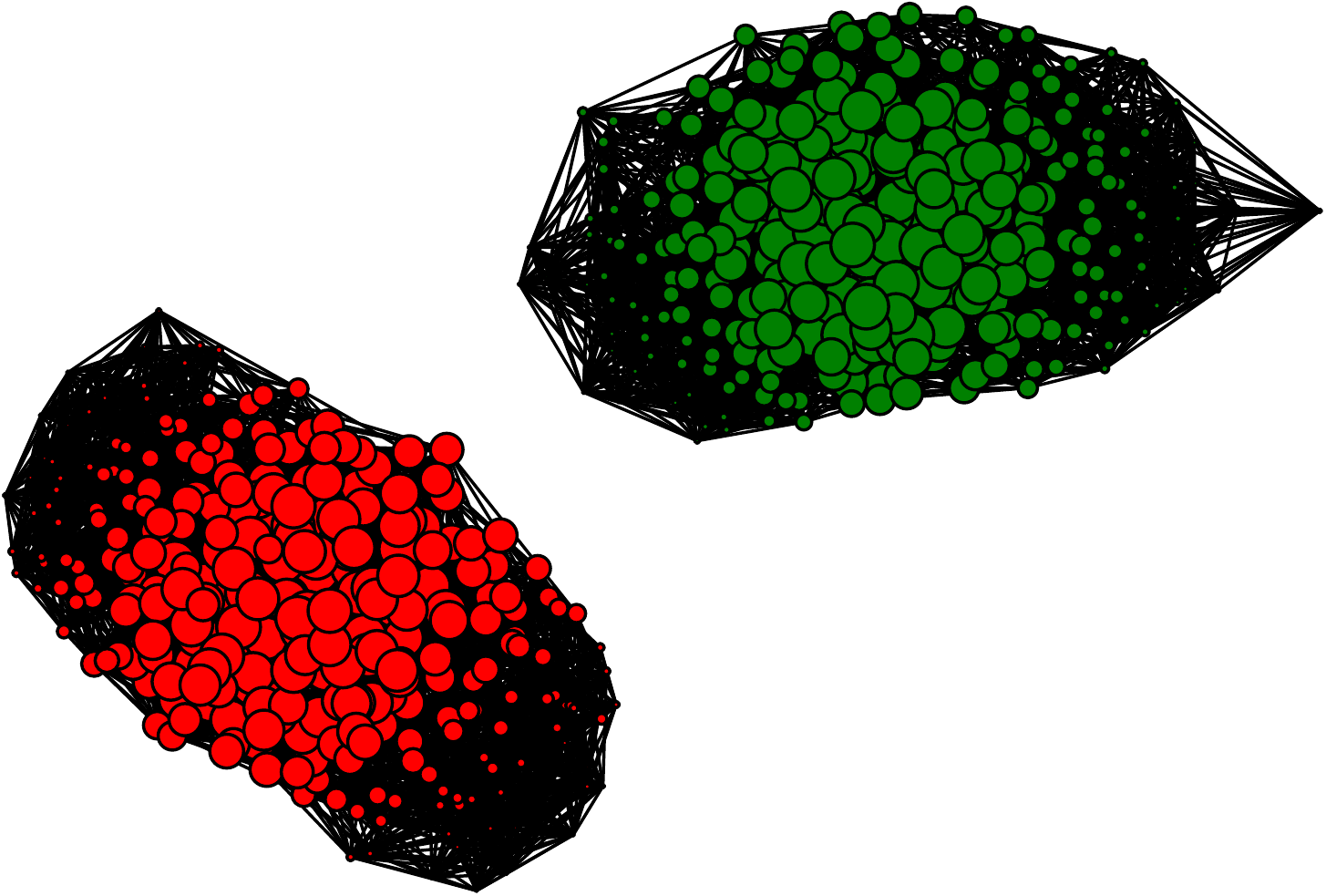}
\label{fig:ds100dim_mv}}
~
\subfigure[Density of differences for each sub-graph.]{
\includegraphics[width=0.33\textwidth,keepaspectratio=true]{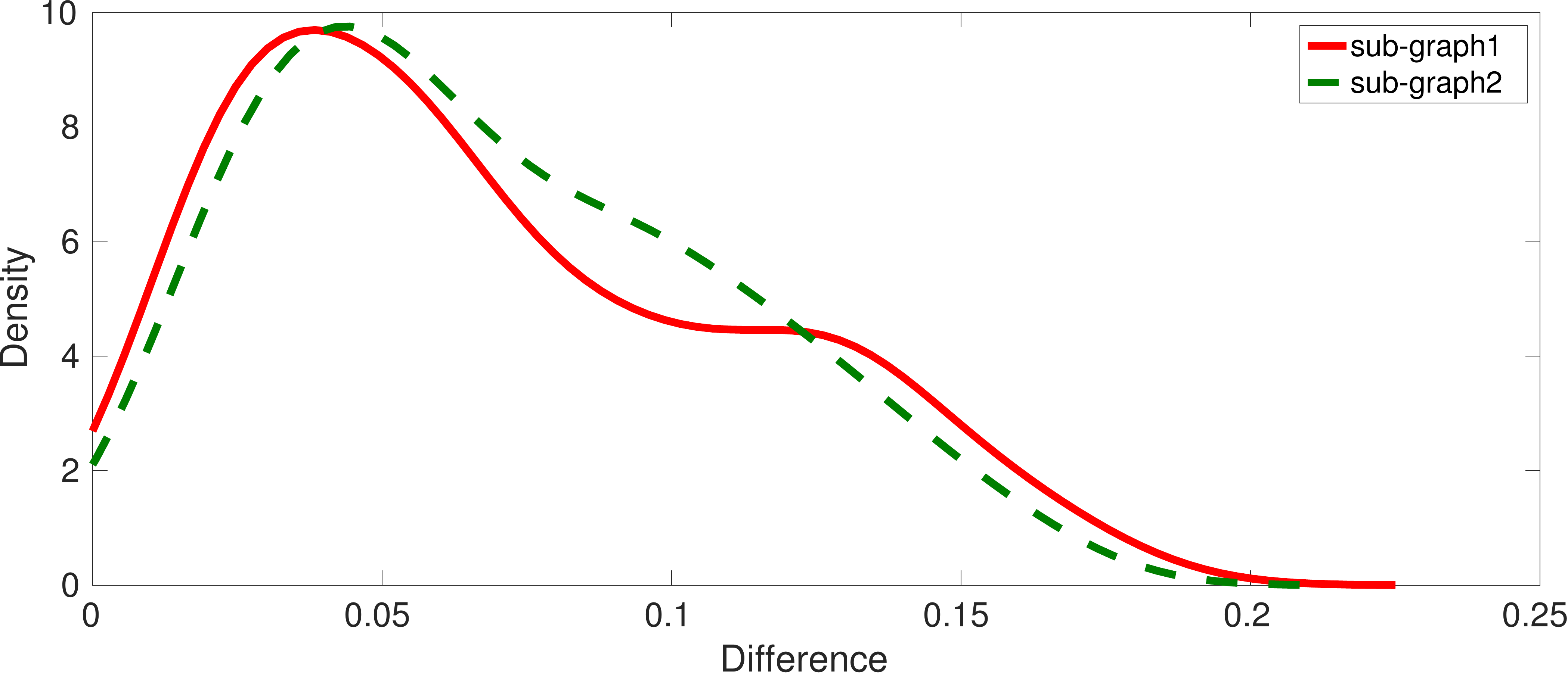}
\label{fig:ds100dim_CCDiff_density}}

\caption{(colors online) Functioning of the graph-based fuzzification on four simple datasets. The graphs are the result of the model training with Alg. \ref{alg:mi_minimization}. Fuzzification is performed on the training set. Size of vertices in the middle panel is proportional to the related membership degree in decision regions. Position of graph vertices in the middle panel plots is not related to the position of the related samples in the input domain. On the right-hand side panels, we show the density of the differences, for each identified sub-graph, with respect to the vertex having maximum closeness centrality; see Eq. \ref{eq:CCDiff}.}
\label{fig:ds_examples}
\end{figure*}

Figure \ref{fig:ds_examples} shows four sample datasets that are useful to highlight key modeling features of the proposed method.
For each dataset (left panels), we show the resulting graph-based model (middle panels) learned during the synthesis; the fuzzification is performed by using the same training data. We show also the density (right panels) of the differences with respect to the vertex with maximum closeness centrality (considering all derived sub-graphs). Such information are useful to understand how the threshold (\ref{eq:binary_decision}) is derived.
The first dataset in Fig. \ref{fig:ds1} shows a simple case with spherical and separated clusters.
The resulting graph-model (Fig. \ref{fig:ds1_mv}) is composed by three decision regions.
The size of each of vertex reflects the membership degree assigned to the corresponding sample by using the herein described graph-based fuzzification mechanism; the edge lengths are monotonically related to the corresponding Euclidean distances. It is easy to recognize the correspondence with the geometry of the dataset.
In the second example (Fig. \ref{fig:ds4}) we show a uniformly distributed dataset. The proposed method learns a graph-based model (Fig. \ref{fig:ds4_mv}) with only one decision region, as in fact the data possess no structure. However, samples still present differences in terms of centrality/membership degree.
In order to stress the capability of the proposed method to model data with intricate geometry, we show in Fig. \ref{fig:ds_crescentfullmoon} a dataset with two clusters having very different geometric properties. The resulting graph-based model shown in Fig. \ref{fig:ds_crescentmoon_mv} correctly finds two decision regions having good resemblance with the geometry of the dataset.
Finally, in Fig. \ref{fig:ds100dim_pca} we show the first two principal components of a 100-dimensional, normally distributed dataset with two well-defined clusters. The graph-model in Fig. \ref{fig:ds100dim_mv} shows that it is possible to perfectly reconstruct such two clusters, even by considering the high-dimensionality of the data.

In all cases, corresponding densities shown on the right-hand side panels (see Figs. \ref{fig:ds1_CCDiff_density}, \ref{fig:ds4_CCDiff_density}, \ref{fig:ds_crescentmoon_CCDiff_density}, and \ref{fig:ds100dim_CCDiff_density}) denote non-trivial distributions of closeness centrality differences (\ref{eq:CCDiff}).
This highlights the need to consider suitable percentiles (\ref{eq:CCDiff_stats}) of such distributions in order to compute the binary decisions of the classifier (\ref{eq:binary_decision})

\subsection{Analysis of computational complexity}
\label{sec:comp_complex}

We assume to perform the global optimization in Algorithm \ref{alg:mi_minimization} with a genetic algorithm.
The computational complexity of the training phase is $O(IPF)$, where $I$ is the maximum number of iterations, $P$ is the population size, and $F$ is the cost of the fitness function.
In the following, we include all relevant terms and factors in the analysis of the computational complexity, even those that do not influence the asymptotic behavior.
The cost of $F$ can be expressed as the sum of the following terms:
\begin{equation}
\label{eq:computational_complexity_training}
\begin{aligned}
O(F_1) &= O(n r D); \\
O(F_2) &= O(\sqrt{n} F_3); \\
O(F_3) &= O(\underbrace{rn^2}_{\mathrm{kNN}} + \underbrace{n}_{\mathrm{components}} + \underbrace{nk}_{\alpha\mathrm{-Jensen}}).
\end{aligned}
\end{equation}

$O(F_1)$ accounts for the construction of the DSR with $n$ samples and $r$ prototypes, where $D$ is the computational complexity associated with the dissimilarity measure for the input data, which, depending on the specific case, might not be negligible.
$O(F_2)$ accounts for the loop at line \ref{alg:global_k}, which repeats at most $\sqrt{n}$ times the operations described by $F_3$.
$O(F_3)$ describes the cost associated with kNN graph construction, determination of the connected components, and computation of $\alpha$-Jensen difference (\ref{eq:alpha_jensen_graph}).
The asymptotic computational complexity for training is hence upper bounded by $O(n^{5/2})$.

Let us now focus on the computational complexity of the test phase (normal operating mode; Sec. \ref{sec:fuzzy_model}).
At this stage, the data are partitioned into $d\geq 1$ components (the decision regions).
Each test sample is sequentially assigned to each of those components. The related kNN sub-graphs are updated with corresponding calculation of closeness centrality and determination of membership degree of the test sample.
The computational complexity is given by:
\begin{equation}
\label{eq:computational_complexity_test}
O\left(mrD\times d\left[\underbrace{r(n/d)^2}_{\mathrm{kNN\ update}} + \underbrace{(n/d)^3}_{\mathrm{centrality}} + \underbrace{n/d}_{\mathrm{membership}} \right] \right).
\end{equation}

In (\ref{eq:computational_complexity_test}), $m$ is the number of test samples.
The first cost, $mrD$, is due to the embedding of the test data in the dissimilarity space.
There are three main costs involved in the computational complexity for testing the model: update of the kNN sub-graph related to a decision region, calculation of closeness centrality for the vertices in the sub-graph, and computation of membership degree of the test sample (see Eqs. \ref{eq:v_closeness_centrality}--\ref{eq:binary_decision}).
The asymptotic computational complexity of the test stage depends thus on the order of $d$. If $d=1$ and assuming $m\ll n$, then the worst-case is $O(n^3)$, which is given by the cost of well-known Floyd-Warshall algorithm for computing all-pairs shortest paths in a weighted undirected graphs.

Asymptotic computational complexity for constructing a kNN graph can be lowered by considering more advanced, yet approximate solutions \cite{chen2009fast,kybic2012approximate}.
Similarly, the vertex centrality can be computed with approximated versions of closeness centrality \cite{borassi2015fast} or by using other measures of centrality (importance) of vertices taken from the complex network literature.
In addition, it is worth noting that, when processing input samples represented as numeric vectors, there is no need to perform the embedding step devised in our method. This would result in a significant reduction of the computational complexity costs (in terms of both time and space) discussed here.

\section{Experimental results}
\label{sec:experiments}

Experimental evaluation is first performed (Sec. \ref{sec:uci_iam_datasets}) on UCI and IAM benchmarking datasets.
The proposed one-class classifier is denoted as EOCC-MI throughout the experiments.
In addition, in Sec. \ref{sec:protein_sol} we address the important problem of protein solubility recognition.
In this case, we take into account and compare several data representation for proteins, including sequences, graphs, and feature vectors.

EOCC-MI depends on two hyper-parameters: the threshold $\tau\geq0$ for terminating the search in Alg. \ref{alg:mi_minimization} and $l$ expressing the $l$th percentile of the distribution of closeness centrality differences (see Sec. \ref{sec:fuzzy_model}).
These two hyper-parameters allow to fine tune the classifier with respect to the problem at hand. However, we noted that, in general, setting $\tau=0.05$ and $l=50$ provides a good performance for all considered problems. Therefore, in the following we adopt such settings.

\subsection{UCI and IAM datasets}
\label{sec:uci_iam_datasets}

\begin{figure*}[ht!]
\centering
\subfigure[Ranking with two-way ANOVA.]{
\includegraphics[width=0.47\textwidth,keepaspectratio=true]{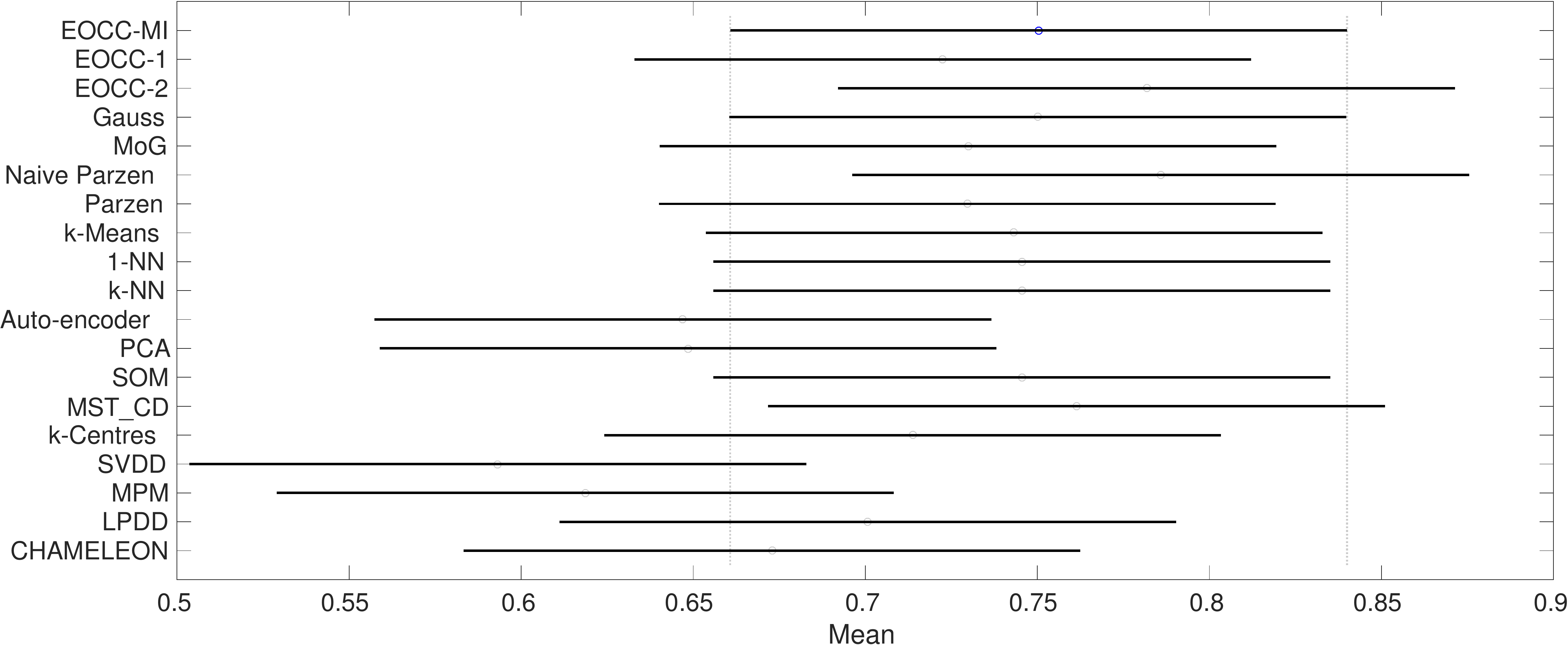}
\label{fig:ANOVA}}
~
\subfigure[Ranking with Friedman test.]{
\includegraphics[width=0.47\textwidth,keepaspectratio=true]{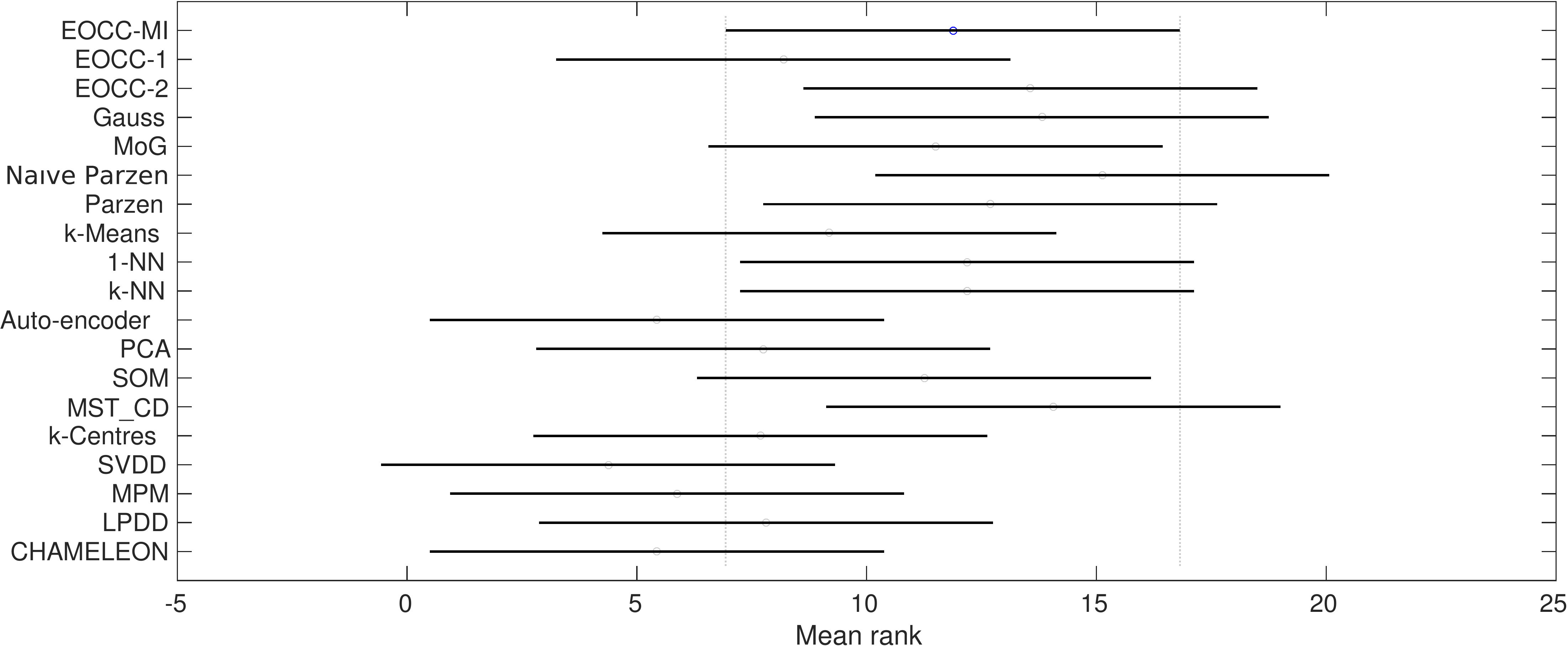}
\label{fig:Friedman}}
\caption{Global ranking of AUC results in Tab. \ref{tab:uci_ds_results}. The proposed classifier, EOCC-MI, denotes competitive performances with respect to the state-of-the-art classifiers taken into account. Missing value for CAMELEON on BC-P is replaced with the mean value over the dataset.}
\label{fig:usi_global_ranking}
\end{figure*}

The UCI datasets taken into account are shown in Tab. \ref{tab:uci_ds}. Datasets and results for comparison are taken from \cite{OCC_results}; datasets are not preprocessed in any way.
AUC results and related standard deviation are shown in Tab. \ref{tab:uci_ds_results}.
EOCC-MI performs well on all UCI dataset with two exceptions: AB and L datasets. In fact, for these two datasets EOCC-MI results are significantly lower than the average outcome ($p<0.0001$).
Global ranking is shown in Fig. \ref{fig:usi_global_ranking}. EOCC-MI denotes global statistics that are not significantly different from the other classifiers taken into account. Considering the two-way ANOVA test (Fig. \ref{fig:ANOVA}), EOCC-MI is globally ranked at the 4th position. When taking into account the Friedman test, instead, it is ranked 8th (Fig. \ref{fig:Friedman}).
\begin{table*}[th!]\scriptsize
\begin{center}
\caption{UCI datasets considered in this study.}
\label{tab:uci_ds}
\begin{tabular}{ccccccc}
\hline
\textbf{UCI Dataset} & \textbf{Acronym} & \textbf{Target class} & \textbf{\# Target} & \textbf{\# Non-target} & \textbf{\# Params} \\
\hline
Abalone & AB & 1 & 1407 & 2770 & 10 \\
Arrhythmia & AR & normal & 237 & 183 & 278 \\
Breast cancer Wisconsin (prognostic) & BC-P & N & 151 & 47 & 33 \\
Biomed & BI & normal & 127 & 67 & 5 \\
Breast Wisconsin & BW & benign & 458 & 241 & 9 \\
Diabetes (prima indians) & D & present & 500 & 268 & 8 \\
Ecoli & E & pp & 52 & 284 & 7 \\
Liver & L & healthy & 200 & 145 & 6 \\
\hline
\end{tabular}
\end{center}
\end{table*}
\begin{table*}[thp!]\scriptsize
\begin{center}
\caption{AUC test set results on UCI datasets.}
\label{tab:uci_ds_results}
\begin{tabular}{ccccccccc}
\hline
&  \textbf{AB}    &   \textbf{AR}     &   \textbf{BC-P}   &    \textbf{BI}    &    \textbf{BW}    &    \textbf{D}     &    \textbf{E}     &  \textbf{L}     \\
\hline
 EOCC-MI   &  0.691(0.012)    &   0.706(0.010)    &   0.569(0.006)    &    0.902(0.007)   &    0.989(0.003)   &    0.712(0.007)   &    0.953(0.008)   &   0.481(0.013)    \\
 EOCC-1    &   0.685(0.013)   &    0.683(0.007)   &    0.554(0.002)   &    0.847(0.002)   &    0.990(0.003)   &    0.607(0.046)   &    0.953(0.002)   &    0.461(0.006)   \\
 EOCC-2    &   0.831(0.001)   &    0.775(0.016)   &    0.585(0.021)   &    0.864(0.003)   &    0.989(0.001)   &    0.717(0.005)   &    0.957(0.003)   &    0.536(0.021)   \\
 Gauss  &    0.861(0.002)   &    0.606(0.006)   &    0.591(0.009)   &    0.900(0.004)   &    0.823(0.002)   &    0.705(0.003)   &    0.929(0.003)   &    0.586(0.005)   \\
 MoG    &    0.853(0.005)   &    0.577(0.166)   &    0.511(0.017)   &    0.912(0.009)   &    0.785(1.003)   &    0.674(0.003)   &    0.920(0.004)   &    0.607(0.006)   \\
 Na\"{\i}ve Parzen &   0.859(0.004)   &    0.774(0.007)   &    0.535(0.015)   &    0.931(0.002)   &    0.965(0.004)   &    0.679(0.003)   &    0.930(0.008)   &    0.614(0.002)   \\
 Parzen    &   0.863(0.001)   &    0.577(0.166)   &    0.586(0.029)   &    0.900(0.011)   &    0.723(0.005)   &    0.676(0.004)   &    0.922(0.004)   &    0.590(0.003)   \\
 \textit{k}-Means  &   0.792(0.011)   &    0.766(0.006)   &    0.536(0.021)   &    0.878(0.012)   &    0.846(0.035)   &    0.659(0.007)   &    0.891(1.006)   &    0.578(1.000)   \\
 1-NN   &    0.865(0.001)   &    0.760(0.008)   &    0.595(0.025)   &    0.891(0.008)   &    0.694(0.006)   &    0.667(0.007)   &    0.902(0.009)   &    0.590(0.009)   \\
 \textit{k}-NN  &    0.865(0.001)   &    0.760(0.008)   &    0.595(0.025)   &    0.891(0.008)   &    0.694(0.006)   &    0.667(0.007)   &    0.902(0.009)   &    0.590(0.009)   \\
 Auto-encoder   &    0.826(0.003)   &    0.522(0.021)   &    0.548(0.037)   &    0.856(0.022)   &    0.384(0.009)   &    0.598(1.008)   &    0.878(1.000)   &    0.564(0.009)   \\
 PCA    &    0.802(0.001)   &    0.807(0.010)   &    0.574(0.018)   &    0.897(0.005)   &    0.303(0.010)   &    0.587(0.002)   &    0.669(0.011)   &    0.549(0.005)   \\
 SOM    &    0.814(0.003)   &    0.772(0.007)   &    0.523(0.030)   &    0.887(0.008)   &    0.790(0.023)   &    0.692(0.007)   &    0.890(0.011)   &    0.596(0.007)   \\
 MST\_CD &    0.875(0.001)   &    0.796(0.006)   &    0.611(0.026)   &    0.898(0.010)   &    0.765(0.018)   &    0.669(0.007)   &    0.897(0.009)   &    0.580(0.009)   \\
 \textit{k}-Centres     &    0.760(0.008)   &    0.767(0.016)   &    0.584(0.055)   &    0.878(0.024)   &    0.715(0.124)   &    0.606(0.016)   &    0.863(0.012)   &    0.537(0.041)   \\
 SVDD   &    0.806(0.001)   &    0.581(0.164)   &    0.498(0.242)   &    0.220(0.003)   &    0.700(0.006)   &    0.577(0.098)   &    0.894(0.008)   &    0.470(0.014)   \\
 MPM    &    0.594(0.001)   &    0.771(0.005)   &    0.053(0.001)   &    0.792(0.057)   &    0.694(0.006)   &    0.656(0.007)   &    0.802(0.005)   &    0.587(0.009)   \\
 LPDD   &    0.697(0.001)   &    0.577(0.166)   &    0.539(0.183)   &    0.865(0.026)   &    0.800(0.005)   &    0.668(0.007)   &    0.896(0.005)   &    0.564(0.026)   \\
 CHAMELEON &   0.706(0.004)   &    0.760(0.008)   &    - &   0.727(0.019)   &    0.669(0.008)   &    0.651(0.010)   &    0.758(0.016)   &    0.580(0.009)   \\
\hline
\end{tabular}
\end{center}
\end{table*}

Tab. \ref{tab:iam_ds} shows some information related to the IAM datasets \cite{riesen+bunke2008}.
The samples (from digital images to proteins) are represented as labeled graphs, with attributes on both vertices and edges.
AUC values are reported in Tab. \ref{tab:iam_ds_results}.
Results are compared with previous experiments \cite{eocc}.
EOCC-MI denotes comparable results with the other two classifiers. However, we note statistically significant differences in two cases (G and P datasets), where EOCC-2 achieves better results.
It is worth underlying that, in EOCC-2 \cite{eocc}, the final model is derived by means of cross-validation. As such, it is expected to be more accurate in terms of recognition capability. However, as shown in the second block of Tab. \ref{tab:iam_ds_results}, EOCC-2 is significantly slower (from tens to hundreds of times slower) in terms computing CPU time required for training (software is written in C++ and implemented on an Intel(R) Core(TM) i7-4710HQ CPU \@2.50GHz).
This fact suggests that EOCC-MI offers a good compromise between computational cost and recognition performance.

\begin{table*}[th!]\scriptsize
\begin{center}
\caption{IAM datasets taken into account.}
\label{tab:iam_ds}
\begin{tabular}{cccccc}
\hline
\textbf{IAM Dataset} & \textbf{Acronym} & \textbf{Nominal class} & \textbf{\# Nominal} & \textbf{\# Non-nominal} \\
\hline
AIDS & A & a & 200 & 1600 \\
GREC & G & 1 & 50 & 1033 \\
Letter-Low & L-L & A & 150 & 1400 \\
Letter-High & L-H & A & 150 & 1400 \\
Protein & P & 1 & 99 & 332 \\
\hline
\end{tabular}
\end{center}
\end{table*}
\begin{table*}[thp!]\scriptsize
\begin{center}
\caption{Average AUC (with related standard deviation) and average serial CPU time (training / test expressed in seconds) for IAM datasets in Tab. \ref{tab:iam_ds}. Results for EOCC-1 and EOCC-2 are taken from \cite{eocc}. Results in bold denote statistically significant differences.}
\label{tab:iam_ds_results}
\begin{tabular}{ccccccc}
\hline
 &  & \textbf{A} & \textbf{G} & \textbf{L-L} & \textbf{L-H} & \textbf{P} \\
\hline
\multirow{3}{*}{\centering\rotatebox{90}{\hspace{-0.7em} \textbf{AUC}}}
 & EOCC-MI & 0.984(0.019) & 0.952(0.008) & 1.000(0.000) & 0.951(0.056) & 0.491(0.018) \\
 & EOCC-1 & 0.977(0.012) & 0.993(0.006) & 1.000(0.000) & 0.905(0.116) & 0.388(0.028) \\
 & EOCC-2 & 0.974(0.014) & \textbf{1.000(0.000)} & 1.000(0.000) & 0.967(0.004) & \textbf{0.554(0.025)} \\
\hline
\multirow{3}{*}{\centering\rotatebox{90}{\hspace{-0.7em} \textbf{CPU}}}
 & EOCC-MI & 14.5/1.922 & 6.376/0.332 & 5.280/0.600 & 8.028/1.462 & 870.1/6.388 \\
 & EOCC-1 & 34.86/0.445 & 6.396/0.168 & 3.901/0.070 & 4.132/0.074 & 568.9/4.867 \\
 & EOCC-2 & 148.2/0.468 & 77.02/0.171 & 151.9/.074 & 168.8/0.093 & 9152/4.955 \\
\hline
\end{tabular}
\end{center}
\end{table*}

\subsection{Protein solubility recognition}
\label{sec:protein_sol}

Protein folding refers to the chemico-physical process transforming the primary structure of a protein into a three-dimensional, active molecular conformation \cite{dill2012protein}. Despite the recent progresses in protein structure characterization and prediction, protein folding is still a largely unsolved problem in biophysics and related computational sciences \cite{Wolynes2014}.
This fact is due to a multitude of causes, such as the large number of residues involved in the process (protein molecules contain from tens to thousands of residues) and the different energy constraints defining the (thermodynamic) energy landscape.
The process of folding strictly competes with the aggregation process, that is, with the tendency of proteins to establish also inter-molecular bonds.
Aggregation propensity is intimately related with the degree of solubility of a molecule \cite{Agostini2012237}.
This results in the formation of large multi-molecular aggregates which, analogously to what happens with artificial polymers, are insoluble and hence precipitate in solution \cite{giuliani2002}. Aggregation propensity of proteins, in turn, is strongly related to problems occurring during the folding process, resulting in pathologies (misfolding diseases) such as Alzheimer and Parkinson \cite{dill2012protein}.
Therefore, studying the solubility degree of proteins is of utmost importance in protein science.

\begin{figure*}[ht!]
\centering
\subfigure[Density of the 3173 E. coli proteins with respect to the normalized solubility degree.]{
   \includegraphics[viewport=0 0 341 243,scale=0.6,keepaspectratio=true]{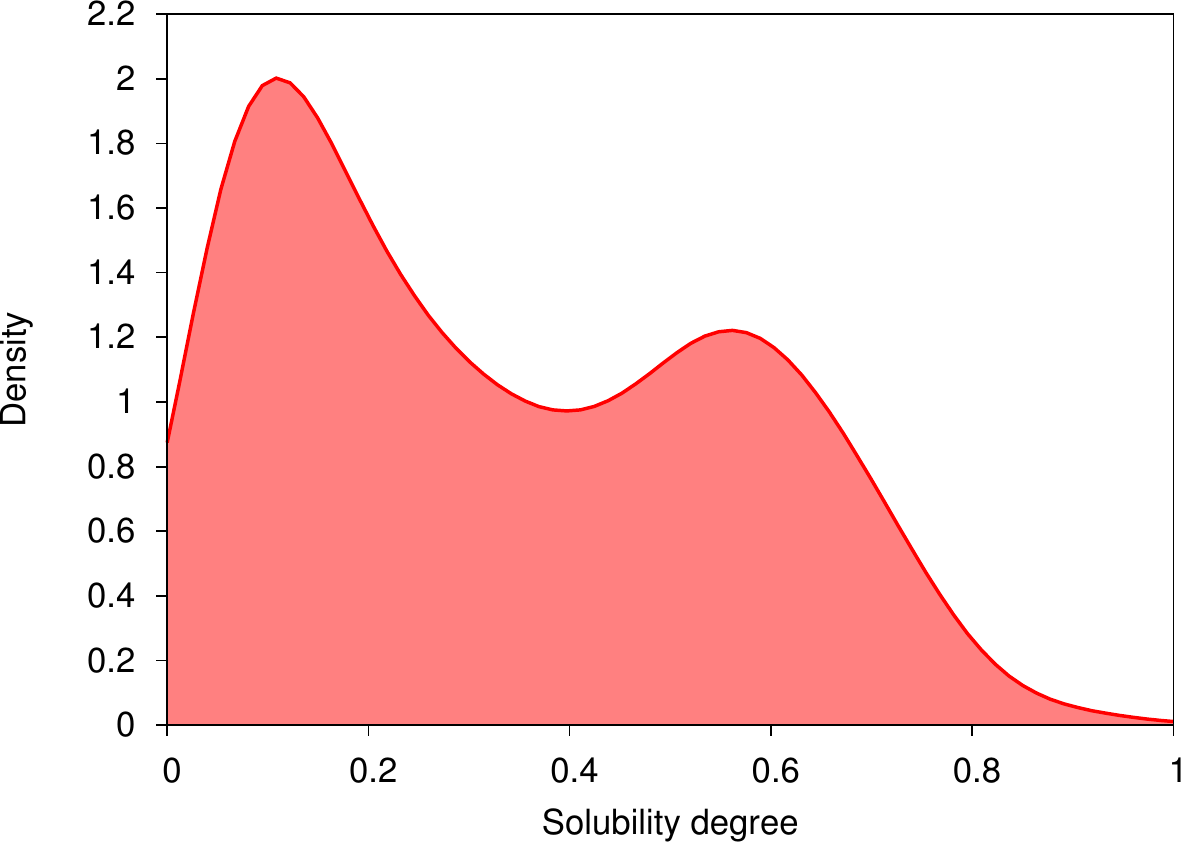}
   \label{fig:solubility}
 }
~
 \subfigure[Normalized solubility degree of the 1811 proteins taken into account. The two classes are clearly recognizable.]{
   \includegraphics[viewport=0 0 351 241,scale=0.6,keepaspectratio=true]{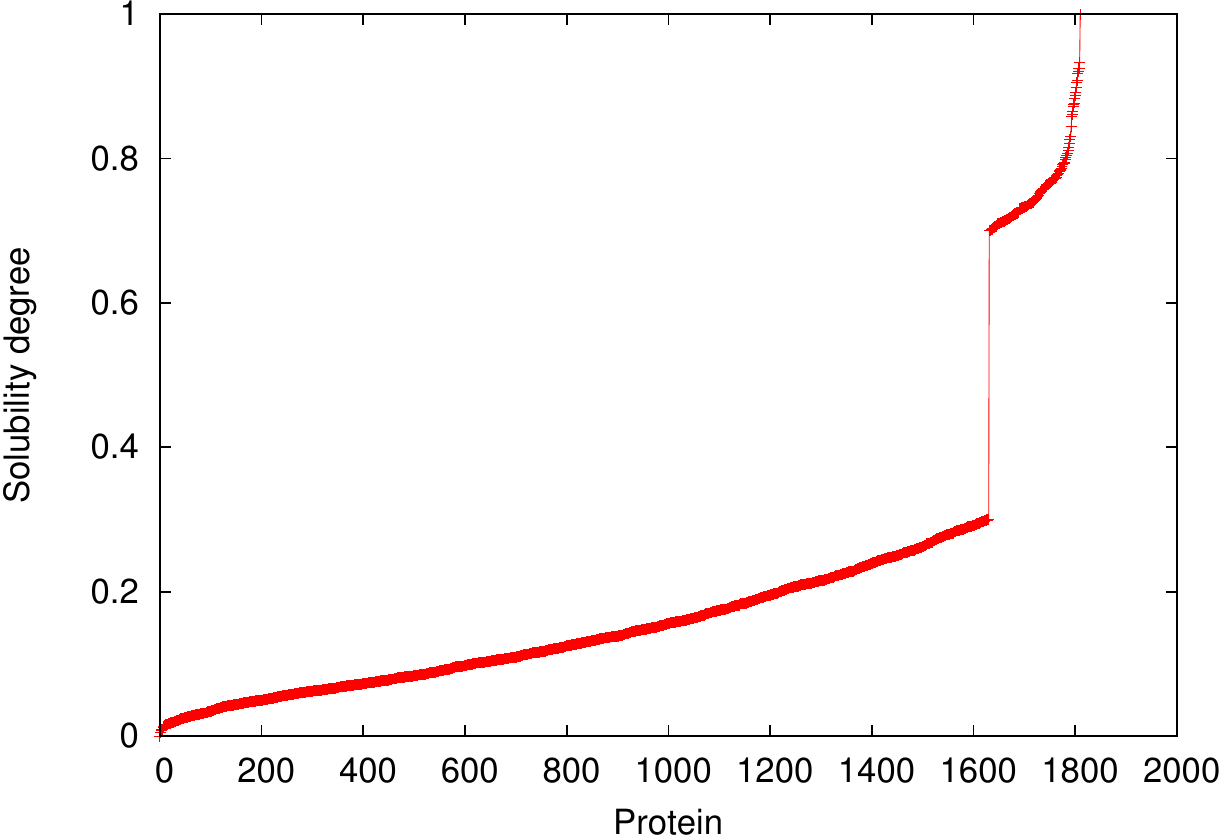}
   \label{fig:norm_sol}
 }
\caption{Density (Fig. \ref{fig:solubility}) and normalized solubility degree (Fig. \ref{fig:norm_sol}) of the E. coli proteins.}
\label{fig:protein_density_sol}
\end{figure*}

Niwa et al. \cite{niwa2009} analyzed in a strictly controlled setting the aggregation/solubility propensity of a large set (3173) of E. coli proteins.
Proteins having difficulty in performing the folding autonomously (i.e., without the help of the so-called chaperones) tend to aggregate and hence precipitate in the solution (water in the experiment).
The 3173 E. coli proteins denote a bi-modal distribution of (normalized) solubility, with many proteins having a low solubility degree and only few very soluble proteins (see Fig. \ref{fig:solubility}).
In order to conceive a classification problem, a suitable threshold for the solubility must be identified within the solubility range.
We consider the $[0, 0.3]$ and $[0.7, 1]$ intervals for determining insoluble and soluble proteins.
The resulting dataset contains 1811 proteins, whose solubility range shown in Fig. \ref{fig:norm_sol} clearly denotes the presence of two different classes: proteins with low and high solubility degree, respectively.
However, although the intervals have the same length and are positioned at the extremes of the solubility range, the resulting dataset is very imbalanced, containing 1631 insoluble and only 180 soluble proteins.

We denote the dataset containing 1811 proteins as DS-1811-SEQ. Samples in this dataset are represented as variable-length sequences of amino acid identifiers (21 different characters that identify the amino acids).
The dissimilarity measure for the input, $d_{\mathrm{I}}(\cdot, \cdot)$, is implemented by using the well-known Levenshtein sequence alignment algorithm.
Folded proteins find a better representation in terms of networks. Therefore, we consider also the corresponding graph-based representation of such proteins.
Vertices of such graphs are labeled with three-dimensional numeric vectors, which convey a compressed version of the information provided by the chemico-physical attributes of the related amino acids. Edges are labeled with the Euclidean distance between the residues.
However, among the 1811 proteins, only 454 currently have a resolved 3D structure in the Protein Data Bank\footnote{http://www.rcsb.org/pdb/home/home.do}.
As a consequence, we were able to generate a dataset of labeled graphs containing 454 proteins. This dataset is denoted as DS-454-GRAPH.
In this case, $d_{\mathrm{I}}(\cdot, \cdot)$ is implemented by using a graph edit distance algorithm \cite{gm_survey}.

\begin{figure*}[ht!]
\centering

\subfigure[PCA.]{
\includegraphics[scale=0.6,keepaspectratio=true]{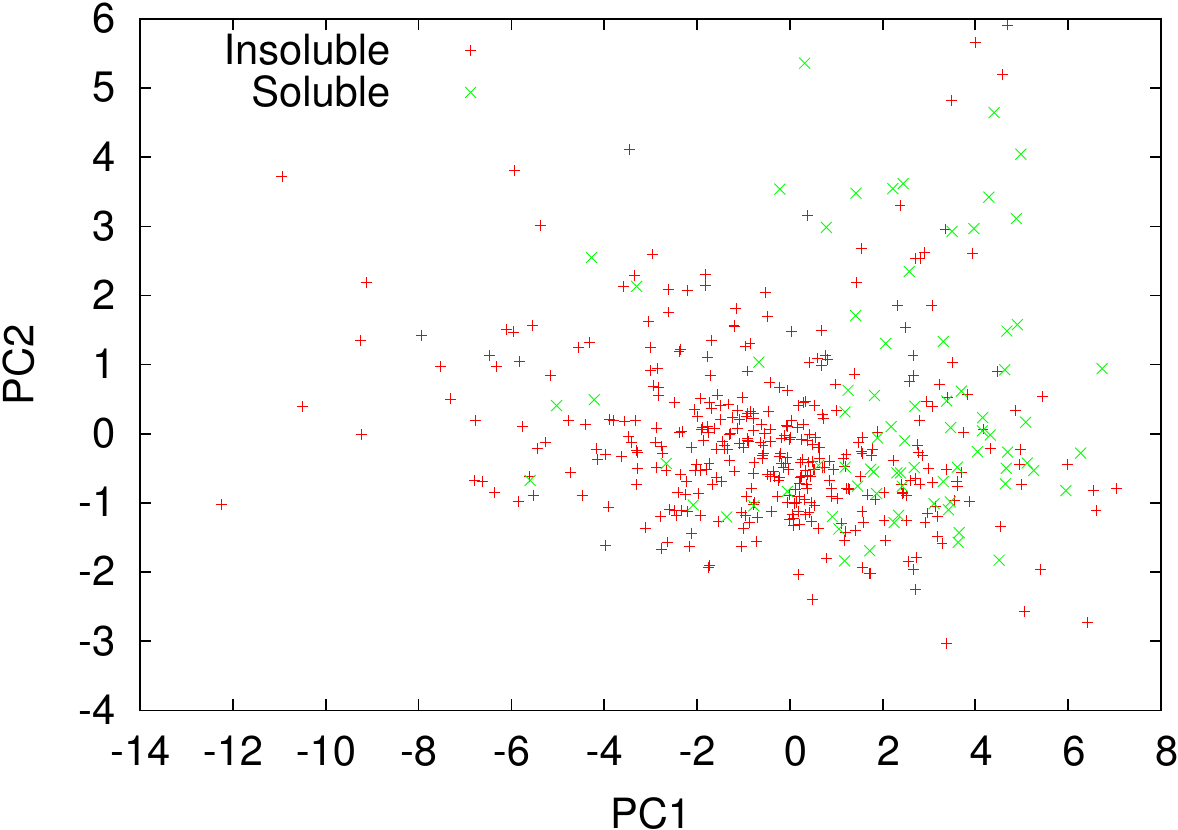}
\label{fig:PCA}}
~
\subfigure[kernel PCA.]{
\includegraphics[scale=0.6,keepaspectratio=true]{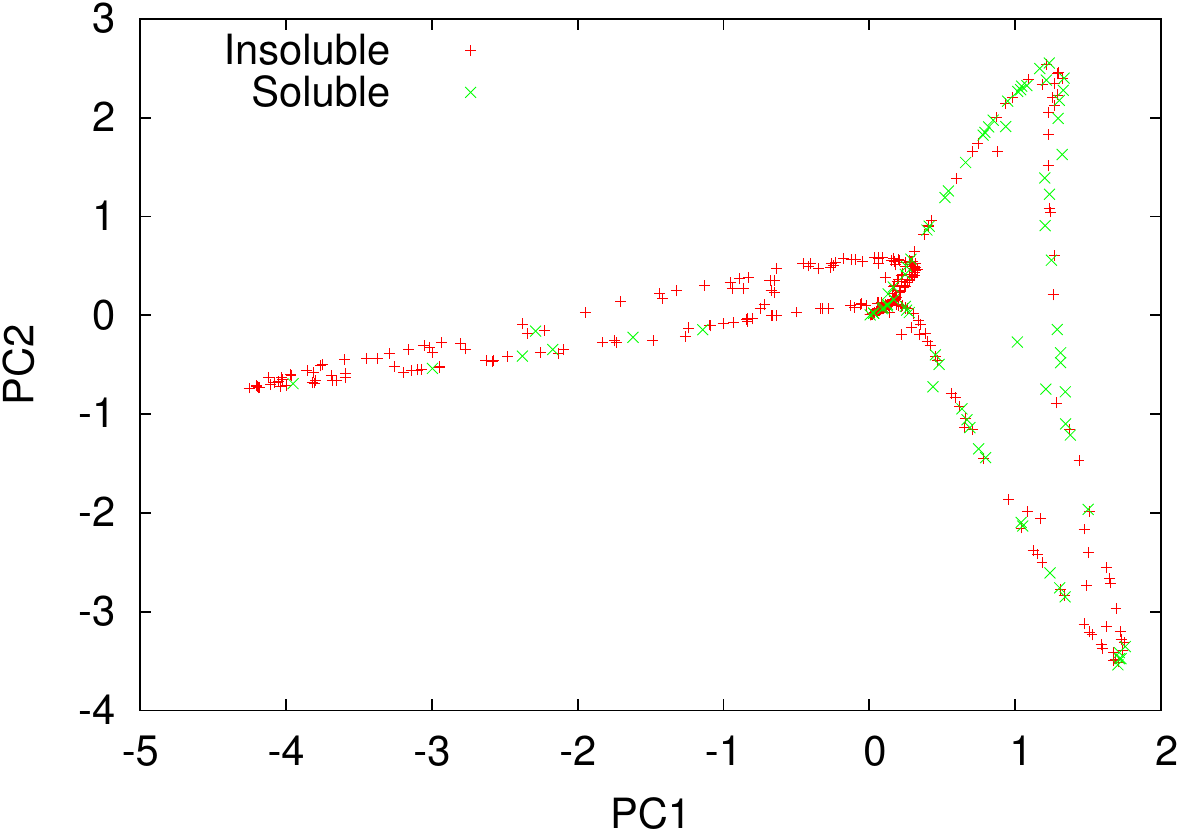}
\label{fig:kPCA}}

\subfigure[Average weights for features with insoluble as nominals.]{
\includegraphics[scale=0.6,keepaspectratio=true]{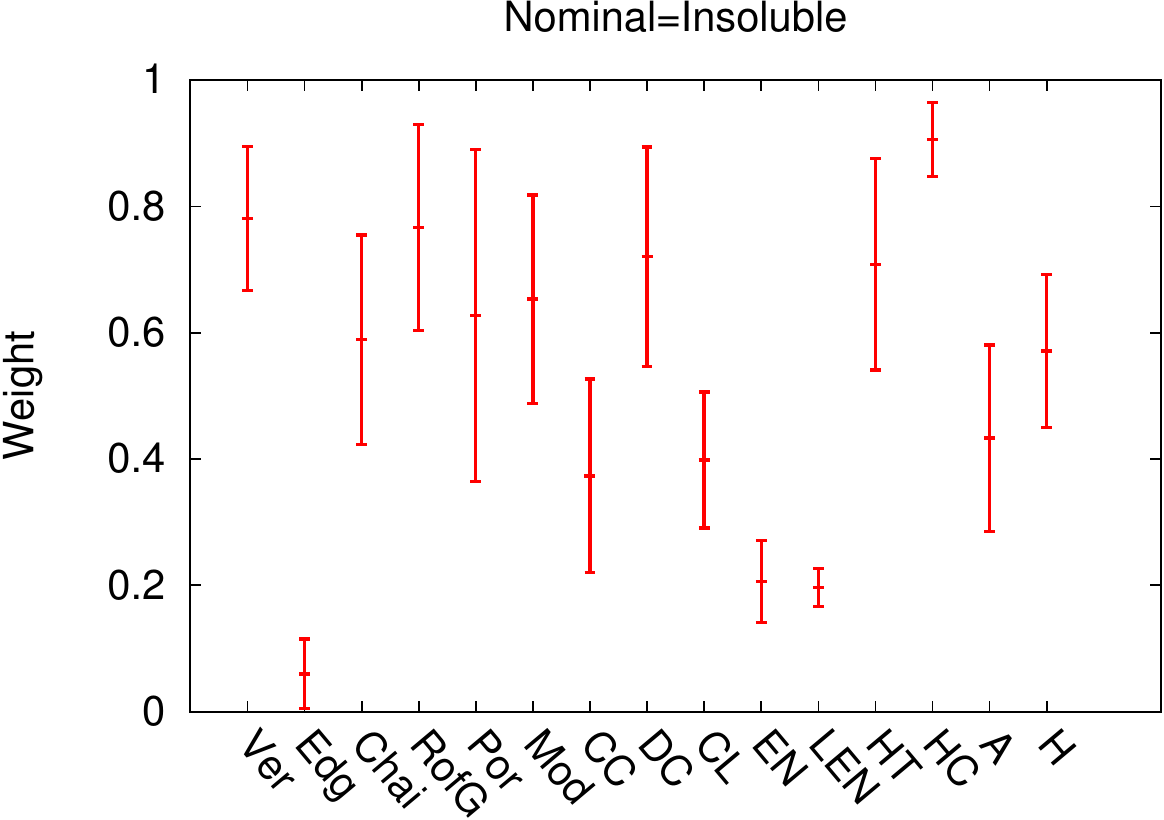}
\label{fig:ecoli_features01}}
~
\subfigure[Average weights for features with soluble as nominals.]{
\includegraphics[scale=0.6,keepaspectratio=true]{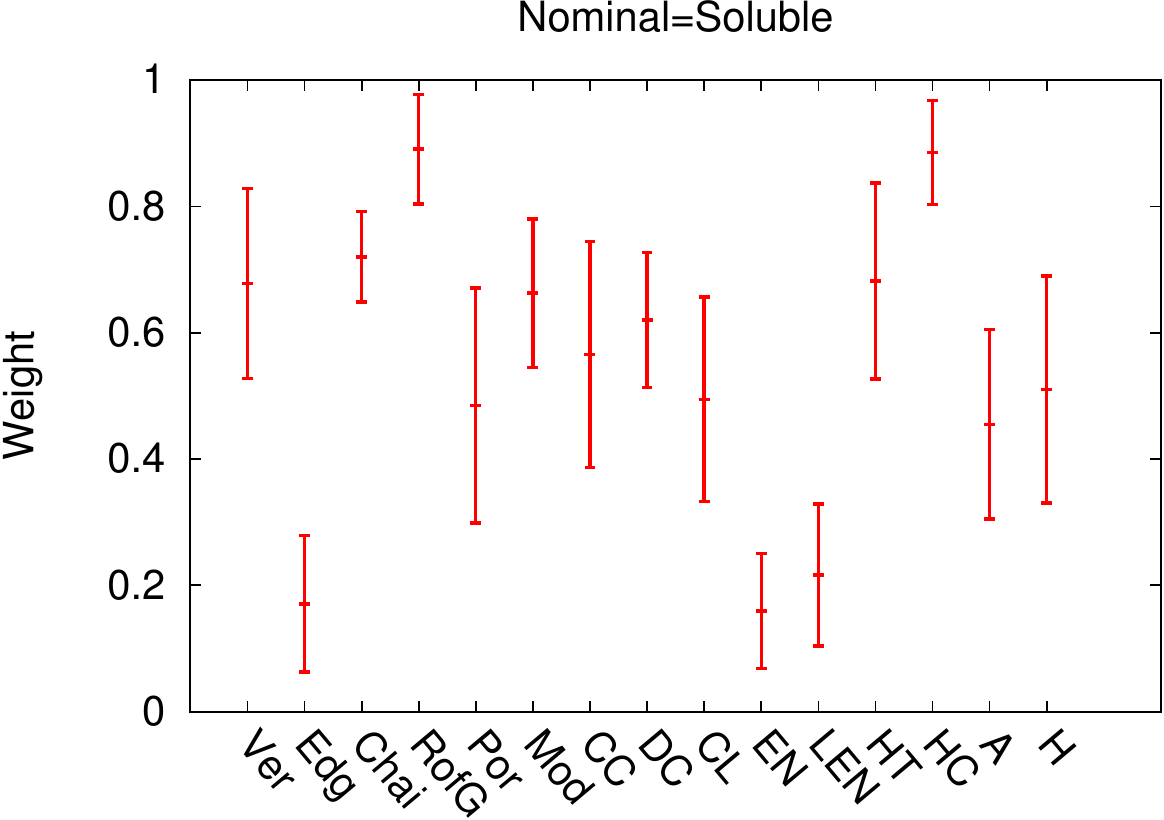}
\label{fig:ecoli_features10}}

\caption{First two components of DS-454-PCA (Fig. \ref{fig:PCA}) and DS-454-KPCA (Fig. \ref{fig:kPCA}). The difficulty of the problem can be appreciated from the high overlap between the two classes. Average weight values learned by the algorithm on DS-454-FEATURE, considering insoluble (Fig. \ref{fig:ecoli_features01}) and soluble proteins (Fig. \ref{fig:ecoli_features10}) for the nominal class.}
\label{fig:ecoli_pca_featurew}
\end{figure*}

By starting from DS-454-GRAPH, we develop four additional data representations.
The first one contains sequences of three-dimensional numeric vectors associated with the graph vertices. This data representation is obtained by ``seriating'' the graphs. Therefore, a sequence contains a number of elements equal to the number of vertices in the related graph; see \cite{ecoli_graph} for details on the seriation algorithm.
In this case, $d_{\mathrm{I}}(\cdot, \cdot)$ is implemented by using the dynamic time warping algorithm, equipped with a weighted Euclidean distance for computing the distances between the three-dimensional vectors composing the sequences.
In addition, we extracted also a collection of numeric features from the graph representations, by considering topological characteristics describing both structural and dynamical features of such graphs.
This consists in mapping each graph in DS-454-GRAPH with a 15-dimensional feature vector. We denote this dataset as DS-454-FEATURE. Finally, in order to reduce dimensionality of DS-454-FEATURE, we post-process such data with principal component analysis and the related kernel version. In the first case, we retain the first five components (explaining $\approx85\%$ of total variance); in the second case, instead, we retain only three components.
The two resulting datasets are denoted as DS-454-PCA and DS-454-KPCA, respectively. First two components of DS-454-PCA and DS-454-KPCA are shown in Figs. \ref{fig:PCA} and \ref{fig:kPCA}, respectively.
In the last three cases, we always use the weighted Euclidean metric for $d_{\mathrm{I}}(\cdot, \cdot)$.

Tab. \ref{tab:proteins_results} shows the AUC results obtained by considering all data representations.
We take into account both cases for defining the class of nominal data, i.e., either the soluble and insoluble proteins.
Results obtained on DS-1811-SEQ are not comparable with the others, due to the different number of samples.
However, related AUC values show that, in both cases, results are comparable with the others.
Let us focus on the remaining five data representations.
When the nominal class is populated with insoluble proteins, significantly better results ($p<0.0001$) are obtained on DS-454-GRAPH.
On the other hand, when considering the soluble proteins as nominal, better results ($p<0.0001$) are obtained on DS-454-SEQV.
This suggests that the rich information provided by the graph representations of proteins is useful for discriminating the degree of protein solubility.

It is worth stressing that results with DS-454-FEATURE are significantly better than those obtained obtained with both DS-454-PCA and DS-454-KPCA.
In fact, AUC is significantly higher ($p<0.0001$ with respect to DS-454-PCA; with DS-454-KPCA, $p<0.0003$ when insoluble proteins are considered as nominal and $p<0.0001$ otherwise). This fact suggests that the selection of the weights $p^{*}$ based on the proposed mutual information minimization criterion results also in an effective mechanism of feature selection.
In Figs. \ref{fig:ecoli_features01} and \ref{fig:ecoli_features10} we show the average values of the weights found by the algorithm for the two scenarios for the nominal class.
Significant features with high average weight are preserved in both cases, yet with some small numerical differences.
The most relevant features are the number of vertices (``Ver''), number of chains in the molecule (``Chai''), radius of gyration (``RofG''), modularity of the graph (``Mod''), average degree (``DC''), and two diffusion characteristics called heat trace (``HT'') and heat content (``HC''), calculated by using Laplacian matrix of the graphs.
\begin{table}[thp!]\scriptsize
\begin{center}
\caption{AUC test set results on all data representations taken into account for the problem of protein solubility recognition.}
\label{tab:proteins_results}
\begin{tabular}{ccc}
\hline
\textbf{Dataset} & \textbf{Nominal=Insoluble} & \textbf{Nominal=Soluble} \\
\hline
DS-1811-SEQ & 0.643(0.015) & 0.765(0.010) \\
\hline
DS-454-GRAPH & 0.764(0.025) & 0.632(0.074) \\
DS-454-SEQV & 0.647(0.033) & 0.783(0.012) \\
DS-454-FEATURE & 0.682(0.026) & 0.759(0.019) \\
DS-454-PCA & 0.611(0.036) & 0.556(0.066) \\
DS-454-KPCA & 0.656(0.014) & 0.584(0.090) \\
\hline
\end{tabular}
\end{center}
\end{table}

\section{Conclusions}
\label{sec:conclusions}

In this paper, we presented a design methodology for one-classifiers based on entropic spanning graphs.
Input data are first mapped into a dissimilarity representation, allowing to deal with several input data representations.
Embedding vectors are constructed by using a (parametric) dissimilarity measure.
An entropic spanning graph is then constructed over the embedded data and successively processed to derive a model for the classifier.
Entropic spanning graphs are well-known in the literature for providing a non-parametric approach to estimate information-theoretic quantities, such as entropy and divergence.
Here, we have proposed to use a k-nearest neighbour graph (a particular instance of entropic spanning graphs) also to define the model of the one-class classifier.
In particular, decision regions have been derived by partitioning the k-nearest neighbour graph vertices considering the related connected components induced during training.
We proposed to guide the selection of the best-performing graph partition by using an optimization criterion based on mutual information minimization.
Notably, we searched for the best-performing parameters $p$ of the input dissimilarity measure and $k$, inducing a partition of order $d$ with minimum statistical dependence.
This was performed to ensure the maximum independence between the different clusters of vertices (i.e., the decision regions forming the model).
Mutual information has been computed by exploiting a convenient formulation defined in terms of the $\alpha$-Jensen difference.

In order to associate a confidence level with each classifier decision, after training we constructed a fuzzy model on the resulting graph partition.
A membership degree is assigned to each vertex of the k-nearest neighbour graph, denoting the membership of the related input sample to the class of nominal data.
The fuzzification of the vertices is based only on the topological information derived from the k-nearest neighbour graph, hence allowing to model datasets with complex geometries.

Experimental evaluation of the proposed one-classifier is performed on both benchmarking datasets (containing samples represented as either feature vectors and labeled graphs) and by facing an application involving protein molecules.
Benchmarking results on numerical data showed that the proposed method performs well with respect to several state-of-the-art one-class classifiers.
A drawback of our approach is the computational complexity, which does not scale well when the sample size grows (it is of the order of $O(n^{5/2})$).
On the other hand, our approach allows to deal with a wide range of problems, regardless of the data representation adopted for the input data.

Here, we tackled the important problem of recognizing the degree of solubility of a datasets of E. coli proteins.
Such data have been analyzed by considering several different data representations, including sequences, labeled graphs, and numeric features.
The possibility to address a given problem from different angles is a valuable design asset characterizing the proposed method.
The analysis of protein solubility revealed the possibility to achieve good performances by considering different data representations, especially with those based on labeled graphs.
In addition, in the case of numeric features used to represent proteins, the proposed method provided a selection of the most relevant ones that could be exploited for further analyses.

Future research efforts will focus on improving the computational complexity, by conceiving approximate or alternative algorithmic solutions for the computation of k-nearest neighbour graph and vertex centrality.
In addition, it could be interesting to conceive a variant of the method able to dealing with time-variant environments. In fact, it is reasonable to think that nominal conditions might change over time, for instance due to concept drifts (e.g., ageing) affecting the data generating process. The one-class classifier should be able to adapt the model on-the-fly with such new nominal conditions, possibly forming new decision regions and/or integrating the already existing ones with new data.

\balance
\bibliographystyle{./IEEEtran}
\bibliography{Bibliography.bib}

\end{document}